%% file: main.tex
\documentclass[10pt,twocolumn,letterpaper]{article}

\usepackage[pagenumbers]{cvpr}
\usepackage[colorinlistoftodos,prependcaption,textsize=tiny]{todonotes}

\input{additional_files/preamble}

\definecolor{cvprblue}{rgb}{0.21,0.49,0.74}
\usepackage[pagebackref,breaklinks,colorlinks,allcolors=cvprblue]{hyperref}

\def\confName{CVPR}
\def\confYear{2027}

\title{Don't Test Where You Trained: Geographic Overlap Inflates End-to-End Driving Scores}

\author{Willi Poh\thanks{Equal contribution.}\\
Karlsruhe Institute of Technology\\
{\tt\small willi.poh@kit.edu}
\and
Kevin Rösch\footnotemark[1]\\
FZI Research Center for Information Technology\\
{\tt\small kevin.roesch@fzi.de}
\and
Christoph Stiller\\
Karlsruhe Institute of Technology\\
{\tt\small christoph.stiller@kit.edu}
}

\begin{document}
\maketitle
\input{sec/0_abstract}
\input{sec/1_introduction}
\input{sec/2_relatedwork}
\input{sec/3_method}
\input{sec/4_results}

\input{sec/5_conclusion}
\input{sec/6_ai_statement}
{
    \small
    \bibliographystyle{ieeenat_fullname}
    \bibliography{main}
}

\input{sec/X_suppl}

\end{document}

%% file: additional_files/preamble.tex
\usepackage{booktabs} %
\usepackage{float}    %
\usepackage{textcomp} %
\usepackage{tabularx} %
\usepackage{multirow} %
\usepackage{colortbl} %
\expandafter\ifx\csname iflatexml\endcsname\relax\expandafter\newif\csname iflatexml\endcsname\fi
\iflatexml
  \newcommand{\legendbox}[1]{\textcolor{#1}{■}}
  \newcommand{\legendline}[2]{\textcolor{#1}{━}}
  \newcommand{\legendoutline}{□}
\else
  \newcommand{\legendbox}[1]{\textcolor{#1}{\rule{0.8em}{0.8em}}}
  \newcommand{\legendline}[2]{\textcolor{#1}{\rule[0.5ex]{0.75em}{#2bp}}}
  \newcommand{\legendoutline}{{\setlength{\fboxsep}{0pt}\setlength{\fboxrule}{0.9pt}\fbox{\rule{0pt}{0.8em}\rule{0.8em}{0pt}}}}
\fi
\definecolor{mplblue}{HTML}{0000FF}
\definecolor{mplred}{HTML}{FF0000}
\definecolor{mplgreen}{HTML}{008000}
\newcolumntype{C}{>{\centering\arraybackslash}X}  %
\newcolumntype{L}{>{\raggedright\arraybackslash}X} %

\usepackage{placeins} %

%% file: sec/0_abstract.tex
\begin{abstract}
In NAVSIMv1, the standard benchmark for end-to-end driving, $89.6\,\%$ of the test scenes lie
within 4\,m of a training scene. We show that this geographic overlap inflates the reported scores.
We propose a geographically disjoint split of NAVSIMv1 and an evaluation that compares
models trained on either split on the same scenes, which neither has seen in training. All
sensor-based models we test, DiffusionDrive, Latent TransFuser, and DrivoR, lose 3.4 to 6.1 PDMS
points on these scenes, mainly by leaving the road and coming close to other objects, against 1.5
points for a baseline without sensors. The loss concentrates
where a training drive of the original split passed the same place. At all other places, models
trained on the original split are no better than those trained on the geographic split.
Models trained on the original split predict road geometry that no camera can see. The gap
persists when the auxiliary map loss is removed, so the models memorize locations even without being
supervised to. On nuScenes, the geographic split cuts SparseDrive's mapping accuracy by $59\,\%$
and also degrades its planning. Driving scores on logged scenes thus partly measure how well a
model remembers the test locations. We release the split to evaluate generalization to new places:
\href{https://github.com/w-poh/geosplit-navsim}{https://github.com/w-poh/geosplit-navsim}.
\end{abstract}

%% file: sec/1_introduction.tex
\section{Introduction}

End-to-end models organize the autonomous driving task in a single model and must
learn from their input to predict safe trajectories in scenarios not seen in training. Such
generalization can only be tested with a valid test set, which must not contain training
data~\cite{kapoor2023leakage}. The driving task adds another constraint: the data must be
geographically disjoint.

\begin{figure}[H]
  \centering
  \includegraphics[width=\linewidth]{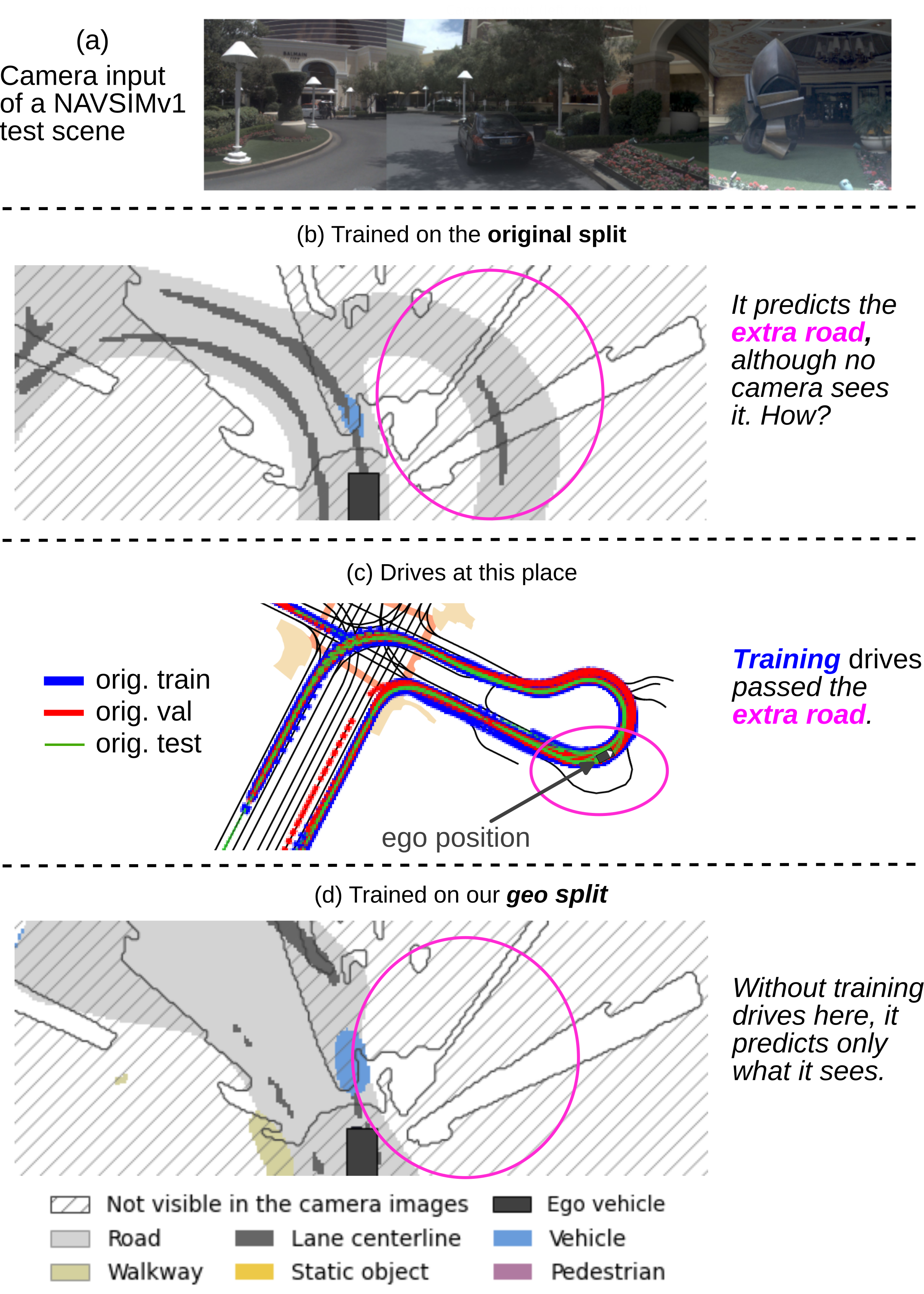}
  \caption{A NAVSIMv1~\cite{navsim} test scene in Las Vegas (a). Trained on the original split, LTF
  predicts a road outside the camera view (b), which training drives passed (c; colors mark the
  train, val, and test sets of the original split). Trained on our geo split, it does not (d).}\label{fig:teaser}
\end{figure}

If the location of a training scene appears in the test set, a model can
memorize the environmental layout, maybe even static instances, and perform better than on
unseen scenes. For online mapping, this has been shown by Yuan \etal~\cite{streammapnet} and Lilja
\etal~\cite{lilja_geosplit}, leading to a reset of benchmark results.
End-to-end benchmarks, however, score only the planned trajectory, yet a model can
still memorize the environment and use it for planning.

 Recent work nevertheless concludes that memorizing locations is an unlikely
explanation for planning performance, based on partly re-rendered scenes~\cite{loewens2026memory}.
We test this on the logged scenes of NAVSIMv1~\cite{navsim}, where $89.6\,\%$ of the test scenes lie
within 4\,m of a training scene, so models can remember the road layout of test places (\cref{fig:teaser}). We evaluate two BEV-based models, an
implicit-representation-based model, and a baseline without visual input on NAVSIMv1, and a modular
model on nuScenes. We show that the overlap inflates the PDMS of all sensor-based models we test,
and propose a geo split for a fair evaluation.
To summarize, our contributions are:

\begin{itemize}
  \item \textbf{Novel geo split for NAVSIMv1:} geographically disjoint training, validation, and test
    sets (overlap $0.2\,\%$ instead of $89.6\,\%$) with all four cities and rare scenarios such as
    roundabouts and drop-off zones in every set. We release the split.
  \item \textbf{Quantifying PDMS inflation:} We compare models trained on either split on the same
    scenes, which neither has seen in training, and show that the geographical overlap inflates
    their scores heavily by 3.4 to 6.1 PDMS points.
  \item \textbf{Location memorization in planning:} The PDMS gap concentrates where a training drive
    passed the same place, mainly as road exits and close approaches to other objects; elsewhere,
    models trained on the original split are no better. This holds within cities, across seeds, and even
    when the map loss is removed.
  \item \textbf{Map memorization:} On NAVSIMv1 and nuScenes, models trained on the original split
    predict map elements outside the camera view; on NAVSIMv1, their map quality there is almost as
    high as on visible ground, while the geo split trained models degrade.
\end{itemize}

%% file: sec/2_relatedwork.tex
\section{Related Work}

\textbf{End-to-end planning datasets.}
For end-to-end planning, the dataset landscape has grown significantly recently
due to the variety of methods and their requirements. At a high level, they can
be categorized into real-world and synthetic datasets. Synthetic datasets use
generated data to provide a variety of scenarios and conditions, especially
designed to cover rare events, dangerous situations, or rare boundary conditions,
e.g., for weather or lighting. A main basis for synthetic datasets is CARLA~\cite{carla_dosovitskiy},
featured for example in Bench2Drive~\cite{bench2drive} and NoCrash~\cite{nocrash}.
Most methods though focus on real-world datasets that provide original driving data
and sensor inputs, such as nuScenes~\cite{nuscenes}, nuPlan~\cite{nuplan}, NAVSIM~\cite{navsim, navsimv2}
and the recently released KITScenes dataset~\cite{kitscenes_richard,kitscenes_royden}.
For end-to-end planning, a de facto standard is to evaluate on the NAVSIMv1 dataset,
e.g.,~\cite{diffusiondrive,drivor,rap,goalflow,world4drive,autovla,vadv2,drivesuprim}. NAVSIMv1
uses nuPlan maps and data, but filters the nuPlan data to obtain a
smaller subset with 103k samples for a \texttt{navtrain} set and 12k samples for a
\texttt{navtest} set. To our knowledge, our work is the first to quantify the geographical
overlap of NAVSIMv1 and to propose a geographically disjoint split for it.

The nuScenes~\cite{nuscenes} benchmark is an open-loop benchmark, originally designed for object
detection and tracking, which is also used for end-to-end planning evaluation. A range of
models are evaluated on nuScenes, e.g.,~\cite{sun_sparsedrive_2025,vad,uniad,genad}. The nuScenes
dataset features approximately 1000 driving scenes from Boston and Singapore. Due to its open-loop nature, the dataset is also used in online mapping benchmarks,
e.g.,~\cite{maptrv2,streammapnet}. Roddick and Cipolla~\cite{roddick2020pon} first split nuScenes
geographically; Yuan \etal~\cite{streammapnet} showed that the standard nuScenes and Argoverse~2
splits overlap heavily (about $85\,\%$ and $54\,\%$) and proposed non-overlapping splits, under which
map models lose about half of their accuracy; Lilja \etal~\cite{lilja_geosplit} analyzed this
leakage and its effect on popular approaches in detail. We
analyze this effect on end-to-end planning, selecting SparseDrive~\cite{sun_sparsedrive_2025}
as an open-loop representative that explicitly trains on a mapping auxiliary task.

Li \etal~\cite{egomlp} show that a model without any visual input, using only the ego status,
performs well on nuScenes. We use their Ego-MLP on NAVSIMv1 as a baseline that cannot recognize a
place.

\textbf{Location memorization in end-to-end driving.}
Löwens \etal~\cite{loewens2026memory} remove Pittsburgh from the NAVSIM training data and find
no drop for DrivoR on the two-stage navhard benchmark (their Tab.~7a), whose second stage re-renders scenes off
the logged path. They conclude that memorizing locations is an unlikely explanation for
end-to-end driving performance. Our setting differs: we evaluate on logged scenes, as navtest and
NAVSIMv1 do, where a model can benefit from having seen the same place in training. Repeating their protocol
on logged scenes, we find drops of about 11 PDMS points on Pittsburgh scenes for DiffusionDrive
and LTF (\cref{sec:app_pittsburgh}). On our geo split,
geographic overlap inflates the scores of all sensor-based models we test, including DrivoR
(\cref{tab:R02-03,fig:nsit_combined}); whether this persists under re-rendering is beyond our
evaluation.

%% file: sec/3_method.tex
\section{Geo Split Design}

\subsection{NAVSIMv1 Geo Split}

Similar to the work of Lilja \etal~\cite{lilja_geosplit} for the
nuScenes dataset, we propose a geo split for the NAVSIMv1 dataset.
In the original NAVSIMv1 split~\cite{navsim}, $89.6\,\%$ of the test scenes lie within 4\,m of
a training scene and $80.4\,\%$ within 4\,m of a validation scene. With an overlap this large,
we cannot just ``use the navtest samples that weren't overlapped with training or validation'',
because then the test set would consist of very few samples. In our geo split, the overlap
drops to $0.20\,\%$ and $0.08\,\%$; the few remaining scenes lie at region borders, where the
4\,m radius reaches across.

\begin{figure}[t]
  \centering
  \includegraphics[width=\linewidth]{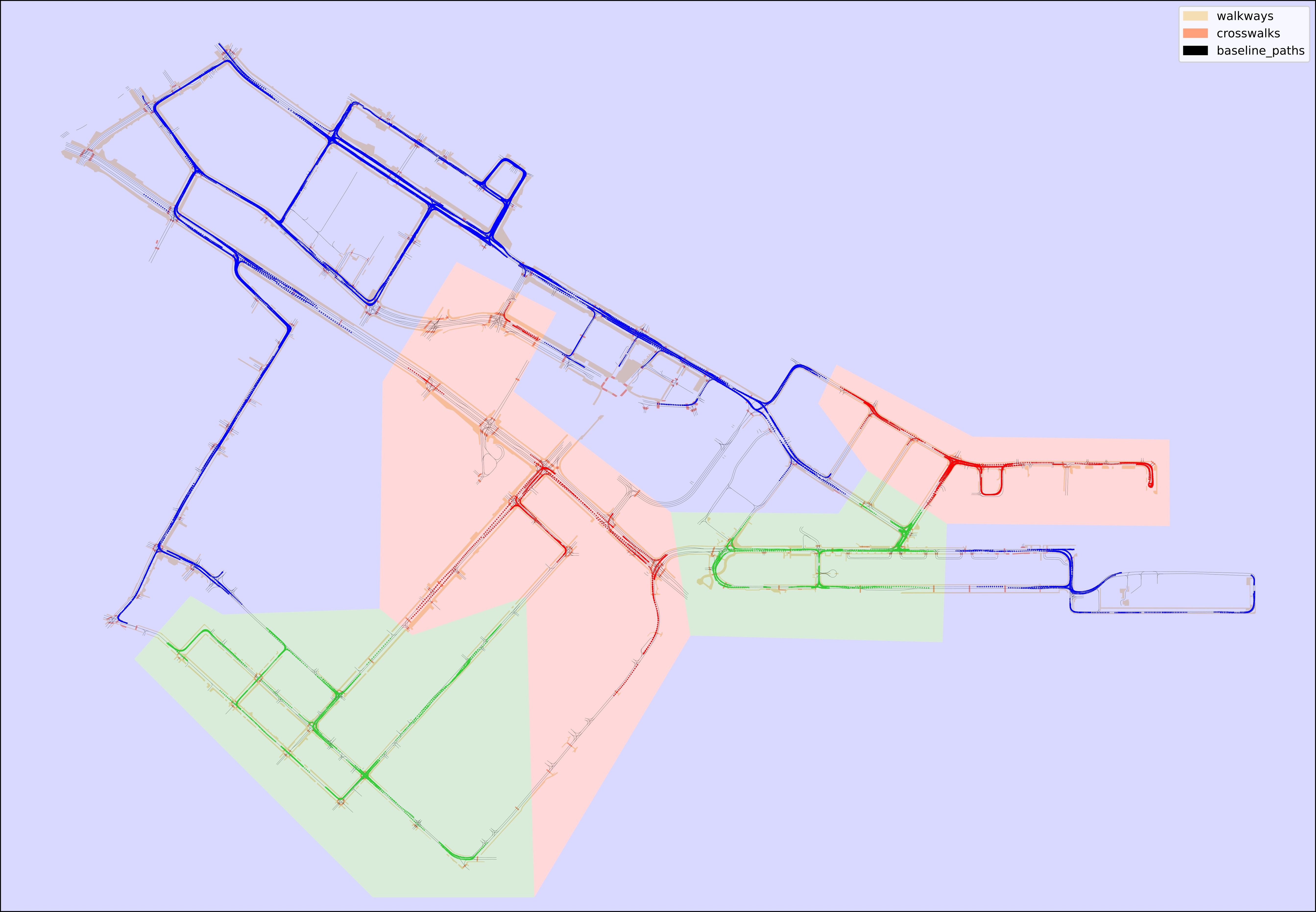}
  \caption{Proposed geographically disjoint split for Boston, US.~The map shows
  the \texttt{val} subset as red polygons (\legendbox{mplred!15}), the \texttt{test}
  subset as green polygons (\legendbox{mplgreen!15}), and the \texttt{train} set as the
  blue background (\legendbox{mplblue!15}).}\label{fig:boston_geo_map}
\end{figure}

For our proposed NAVSIMv1 geo split design, we keep the ratio $r$ of scene shares
$r_\text{train}:r_\text{val}:r_\text{test} = 0.74:0.16:0.11$
similar to the original split. Each of the four cities is represented in
all three subsets. An example for our new Boston geo split can be found in~\cref{fig:boston_geo_map}; the other cities are shown in~\cref{sec:app_maps}.
In our split, we distribute main driving scenarios such as intersections, highways,
roundabouts, left and right turns, and drop-off zones across all sets.

Because whole geographic regions are assigned to one set, the city composition of the sets
changes (\cref{tab:city_composition}): the geo training set contains about 50\,\% more Singapore
and 40\,\% more Boston scenes than the original one, and Las Vegas makes up a larger share of the
geo test set ($55.6\,\%$ vs.\ $33.5\,\%$).

The split is transferable to NAVSIMv2 and nuPlan, which share the same maps.
NAVSIMv2's second stage consists of Gaussian-splatting re-renderings that exist only for
the released navhard logs, at about 1--2\,h of reconstruction per scene~\cite{navsimv2}, so
a geographic test set would need new renderings. NAVSIMv2 scores appear only in our repetition of the
Pittsburgh hold-out of Löwens \etal~\cite{loewens2026memory}, on navhard's real first-stage scenes (\cref{sec:app_pittsburgh}).

\input{sec/tab/tab_city_composition}

\subsection{nuScenes Geo Split}

We follow Lilja \etal~\cite{lilja_geosplit} for the nuScenes geo split, with one additional
filter: scenes that span two sets are split into two scenes, and parts shorter than two frames
are discarded. This drops a small number of scenes while keeping the sets geographically
disjoint. Since planned trajectories can extend beyond scene borders, a small overlap at set
borders remains possible; its impact is negligible given the total number of scenes.

%% file: sec/tab/tab_city_composition.tex
\begin{table}[t]
  \caption{City composition of the NAVSIMv1 sets in the original (Orig.) and our geo split (Geo),
  in \% of the scenes of each set. Scene counts per set in the last row.}\label{tab:city_composition}
  \centering
  \small
  \setlength{\tabcolsep}{3.5pt}
  \begin{tabular*}{\linewidth}{@{\extracolsep{\fill}}lcccccc@{}}
    \toprule
     & \multicolumn{2}{c}{Train} & \multicolumn{2}{c}{Val} & \multicolumn{2}{c}{Test} \\
    \cmidrule(lr){2-3} \cmidrule(lr){4-5} \cmidrule(l){6-7}
    City & Orig. & Geo & Orig. & Geo & Orig. & Geo \\
    \midrule
    Singapore  & 6.1 & 9.4 & 36.4 & 22.2 & 15.8 & 14.2 \\
    Boston     & 10.8 & 15.3 & 20.8 & 10.1 & 30.7 & 14.8 \\
    Pittsburgh & 9.2 & 8.7 & 16.4 & 21.8 & 20.1 & 15.3 \\
    Las Vegas  & 73.9 & 66.6 & 26.4 & 45.9 & 33.5 & 55.6 \\
    \midrule
    {\footnotesize Scenes} & {\footnotesize 85,109} & {\footnotesize 83,945} & {\footnotesize 18,179} & {\footnotesize 15,521} & {\footnotesize 12,146} & {\footnotesize 15,227} \\
    \bottomrule
  \end{tabular*}
\end{table}

%% file: sec/4_results.tex
\section{Results}

\subsection{Experimental Setup}

\textbf{Models.} We evaluate four planners on NAVSIMv1~\cite{navsim}. DiffusionDrive~\cite{diffusiondrive}
uses camera and LiDAR input on a TransFuser backbone and adds a diffusion-based planner. Latent
TransFuser (LTF)~\cite{ltf,navsim} is the camera-only TransFuser variant defined in NAVSIM, in which the
LiDAR input is replaced by a learned latent. Both are trained with an auxiliary map task and encode
spatial information in a BEV representation. DrivoR~\cite{drivor}, a recent leader on the NAVSIM
benchmarks, uses the four cameras only and has no map task and no explicit BEV or map
representation. Ego-MLP~\cite{egomlp} uses only the ego status and the past trajectory, without any
perception, and is a baseline that cannot recognize a place. We report NAVSIMv1's PDM score
(PDMS) and, for DiffusionDrive and LTF, the per-class intersection over union (IoU) of the maps predicted from their BEV features.

We compare each pair of models under two evaluation schemes (\cref{tab:evaluation_schemes}). In
\textit{Original vs.\ Geo} (OVG), each model is evaluated on its own test set, as a benchmark would.
The two test sets differ, so OVG does not compare the models on the same data. In \textit{Never Seen
In Training} (NSIT), both models are evaluated on the same 7,941 samples that neither was trained on;
the original-split model may still have seen their places in other training drives, the geo split
trained model has not. Every NSIT sample lies in the test set of at least one of the two splits
(\cref{tab:evaluation_schemes}); validation sets are not used for model selection. We use the 3,859 samples
in both validation sets as an additional check (\cref{sec:app_valval}).

\input{sec/tab/tab_eval_schemes} %

\textbf{Statistics.} NAVSIMv1 results are averaged over three training seeds for DiffusionDrive
and LTF and one for DrivoR and Ego-MLP (\cref{sec:app_seeds}). We report 95\,\% intervals from a
bootstrap~\cite{efron1992bootstrap} over whole drives (2,000 resamples), since scenes of the same
drive are correlated~\cite{miller2024adding}; for
differences between two models, both are resampled on the same drives.

\textbf{nuScenes.} On nuScenes~\cite{nuscenes}, we use the geo split of Lilja
\etal~\cite{lilja_geosplit} and evaluate SparseDrive~\cite{sun_sparsedrive_2025}, a modular
end-to-end model with auxiliary mapping and detection tasks. Unlike the NAVSIMv1 models, it
represents map elements explicitly as polylines.

\subsection{Main Results of the Geo Split}

\input{sec/tab/R02-03} %

\textbf{PDMS results.} \Cref{tab:R02-03} reports both schemes. On the same scenes (NSIT), all
sensor-based models drop by 3.4 to 6.1 points, DiffusionDrive the most, while Ego-MLP drops by
only 1.5 points, which bounds the effect of the different training sets on a model without
perception. DrivoR drops almost as much as LTF, although it has no map task and no explicit map
representation, so it too seems to encode and use information about the place. On 3,859 further
scenes in the validation set of both splits, the gap persists (DiffusionDrive $+8.7$, LTF $+4.8$;
\cref{sec:app_valval}). In OVG, Ego-MLP even improves on the geo split. We attribute this to the
different composition of the two splits: about two thirds of the geo test set stem from navtrain
drives, and the geo training set contains more scenes from Singapore, the only city with left-hand
traffic (\cref{tab:city_composition}).

\begin{figure*}[t]
  \centering
  \includegraphics[width=0.85\textwidth]{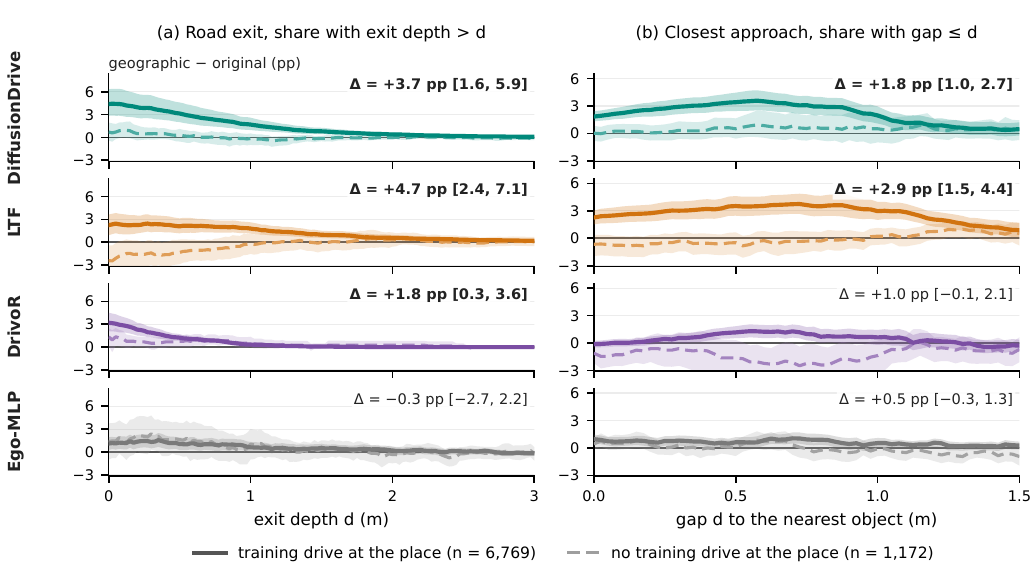}
  \caption{Gap between the geo and the original split trained model on NSIT scenes, with (solid,
  $n = 6{,}769$) and without (dashed, $n = 1{,}172$) an original-split training drive at the place;
  positive: the geo split trained model fails more often. (a)~Share of scenes whose plan leaves the
  drivable area by more than $d$ ($d = 0$: DAC failure). (b)~Share of scenes whose plan comes within
  $d$ of another object ($d = 0$: box overlap). $\Delta$: solid minus dashed at $d = 0$ with 95\,\%
  interval, bold if it excludes zero. Bands: 95\,\% intervals.}\label{fig:nsit_combined}
\end{figure*}

\textbf{Where the gap comes from.} We split the NSIT scenes by whether an original-split training
drive passed within 5\,m in the same direction (6,769 scenes) or not (1,172). The PDMS gap
concentrates in the first group, e.g., for DiffusionDrive $+6.9$ vs.\ $+0.7$ PDMS points
(\cref{sec:app_nsit_drive}); we call this difference the \emph{training-drive effect}. \Cref{fig:nsit_combined} shows which failures cause it. Where no training drive passed (dashed), the curves of
DiffusionDrive and LTF stay near zero, so there the geo split trained model leaves the road and comes
close to other objects about as often as the original-split model. Where a training drive passed
(solid), it does both more often, by 1.8 to 4.7 percentage points more than at unvisited places. It
also leaves the road by more than 1\,m and passes within 0.5\,m of
other objects more often. DrivoR shows the same for road exits, but only weakly for the closest
approach. For Ego-MLP, which cannot recognize a place, solid and dashed curves coincide. The original
split thus does not make the models drive better in general, but better at the places they have seen.

\input{sec/tab/R10}

\input{sec/tab/R38}

\textbf{PDMS analysis.} Most of the gap comes from scenes that fail completely (\cref{tab:R10}).
On the geo split, the share of scenes with PDMS $= 0$ almost doubles for DiffusionDrive and DrivoR,
while the remaining scenes lose at most 1.4 points. For Ego-MLP, the share barely changes. \Cref{tab:R38} splits the gap by failure cause. Road exits of more than $0.5$\,m contribute most
for DiffusionDrive, almost half of its gap ($2.73$ of $6.01$ points), and clearly for LTF ($1.59$),
both beyond the $1.13$ of the sensor-free Ego-MLP, as expected once the
memorized map is no longer available. DrivoR's gap comes mainly from shallow road exits of less
than $0.5$\,m ($1.76$ points). Collisions on the road add about $1.4$ points for DiffusionDrive and LTF, against
$0.5$ for Ego-MLP, and time-to-collision violations without a collision up to $0.8$ points. More
scenes in which the geo split trained models leave the road are shown in \cref{sec:app_qualitative}.

\textbf{Different cities.} Within each city (\cref{tab:R16}), the training-drive effect is positive
for all three sensor-based models in Singapore, Boston, and Pittsburgh, except for DrivoR in
Boston; Ego-MLP shows it in no city, and in Boston its effect is even negative. The effect is therefore not caused by the different city
composition of the two splits. One result goes the other way. Without a training drive, the geo
split trained LTF even scores higher ($-2.8$ PDMS points). This is confined to Singapore and about
ten drives of one vehicle, without which it vanishes ($-0.2$). A plausible cause is the larger
Singapore share of the geo training set (\cref{tab:city_composition}), but a direct test is
inconclusive.

\begin{figure*}[t]
  \centering
  \includegraphics[width=\textwidth]{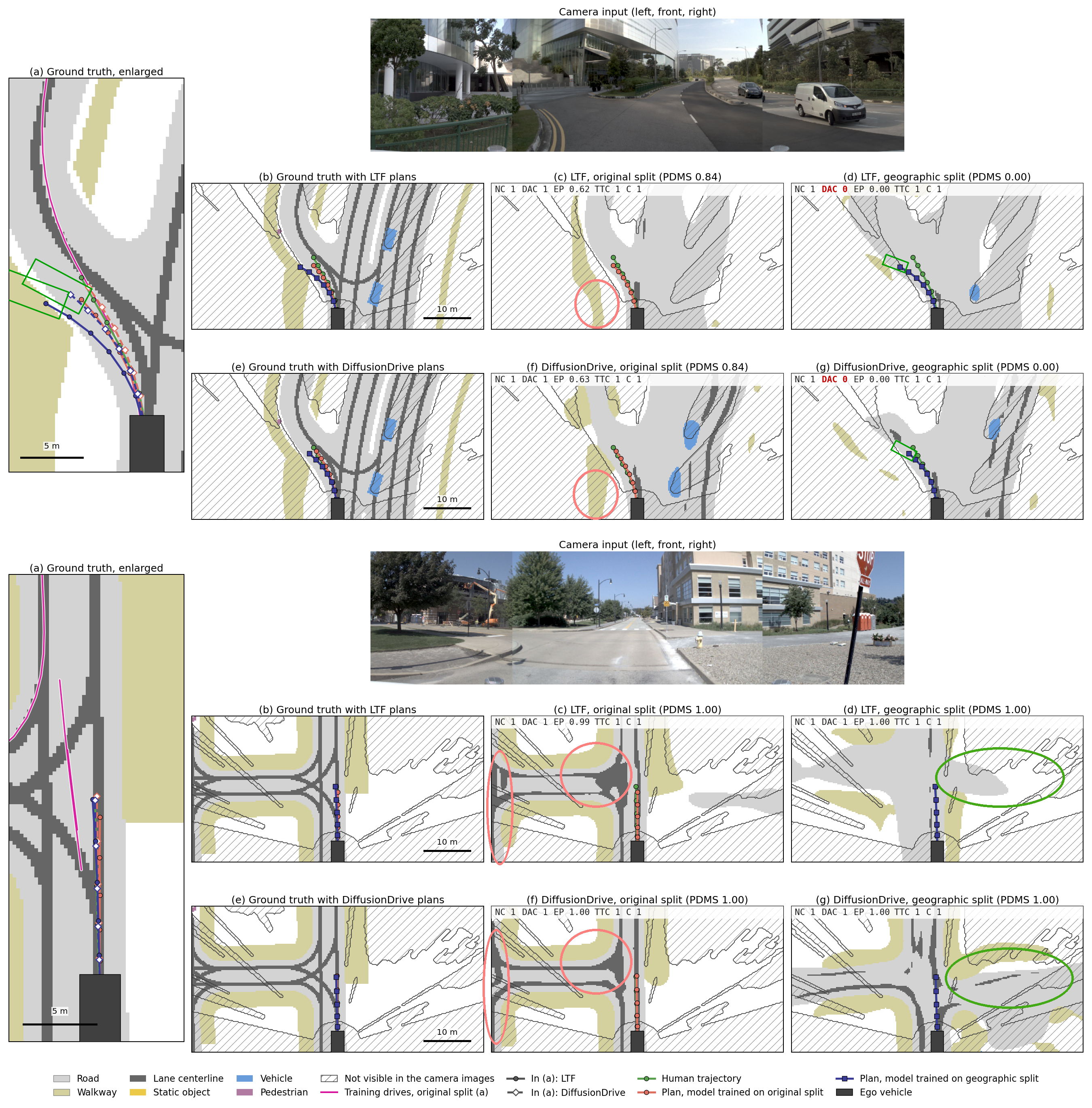}
  \caption{Two NAVSIMv1~\cite{navsim} test scenes. (a)~Ground truth with the training drives of the
  original split within 60\,m of the scene (pink); (b,~e)~ground truth with the plans of LTF and DiffusionDrive; (c,~d,~f,~g)~BEV
  maps predicted by the original-split and the geo split trained models, with PDMS and subscores.
  Red circles: predicted map elements outside the camera view. Green circles: drivable areas that
  the geo split trained models predict, but the ground truth lacks. Green boxes: ego vehicle at the
  step farthest outside the drivable area.}\label{fig:navsim_examples}
\end{figure*}

\textbf{Holding out a city.} A second design gives the same picture. Following Löwens
\etal~\cite{loewens2026memory}, we train DiffusionDrive and LTF on the original split without
Pittsburgh. On navtest, they lose about 11 PDMS points on Pittsburgh scenes ($+10.8$ and $+10.6$),
but less than one point in the other cities; the sensor-free Ego-MLP loses $3.3$ points on
Pittsburgh (\cref{sec:app_pittsburgh}).

\input{sec/tab/R16}

\input{sec/tab/R29}

\textbf{Memorization without map supervision.} Is the auxiliary map loss the channel of the leakage?
We train LTF and DiffusionDrive without it on both splits (\cref{tab:R29}). The gap does not close.
For DiffusionDrive, it stays at $5.5$ points, and for LTF, it even widens to $5.8$ points. The
models thus memorize locations without being supervised to. For LTF, the wider gap comes from the geo split trained model, which performs worse
without the map loss. We suspect that the auxiliary map task
regularizes the whole model through the self-attention between the BEV and the image branch.
Removing the map loss also lowers the absolute PDMS of the original-split LTF (by $2.0$ points,
\cref{sec:app_mapablation}), in line with the TransFuser ablation of NAVSIM~\cite{navsim} (their
Tab.~2); unlike prior ablations, we measure its effect on the gap between the two splits.

\input{sec/tab/R23} %

\input{sec/tab/spdr}

\begin{figure*}[t]
  \centering
  \begin{subfigure}{\textwidth}
    \centering
    \includegraphics[width=\linewidth]{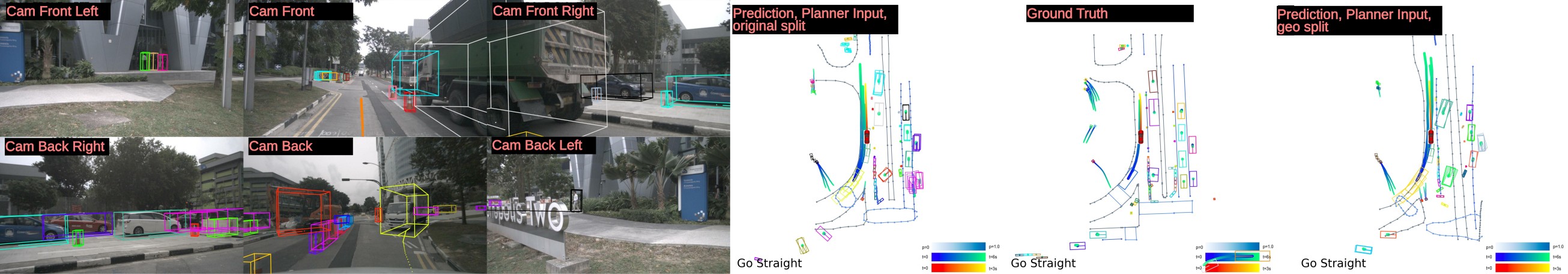}
    \caption{Proceeding straight: the drop-off zone is not visible to the cameras.}\label{fig:curbside_straight}
  \end{subfigure}
  \par\medskip
  \begin{subfigure}{\textwidth}
    \centering
    \includegraphics[width=\linewidth]{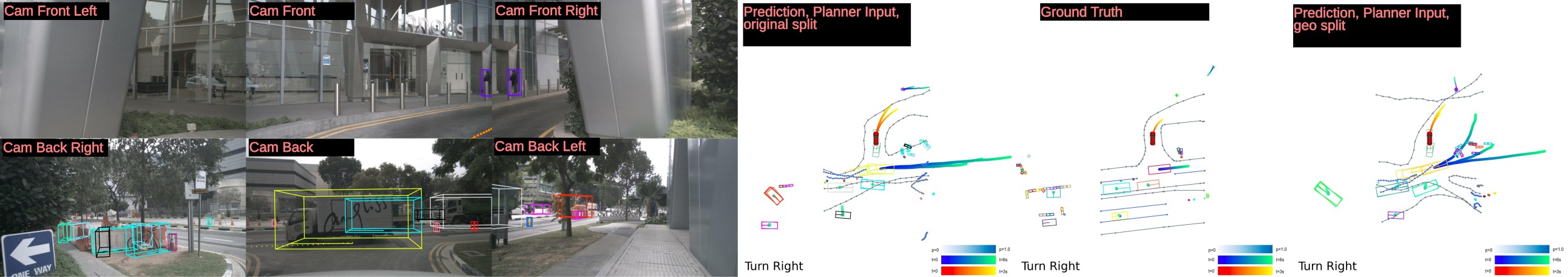}
    \caption{Turning towards the drop-off zone.}\label{fig:curbside_turn}
  \end{subfigure}
  \caption{SparseDrive on a nuScenes~\cite{nuscenes} scene in Singapore with a drop-off zone, trained on the
    original split (top row of each subfigure) and on the geo split~\cite{lilja_geosplit}
    (bottom row). Each row shows the camera images with detections (left), the predicted map,
    agents, and plan used as planner input (middle), and the ground truth (right).}\label{fig:curbside}
\end{figure*}

\textbf{Rankings.} The four models rank the same on both splits, but the distances between them
change: on NSIT, DiffusionDrive's lead over LTF shrinks from $6.4$ to $4.2$ points, while DrivoR's
lead over DiffusionDrive grows from $5.7$ to $8.3$ points. Differences between methods on the
original split therefore partly reflect how much each method profits from memorized locations, and
absolute scores overstate how well the models drive at new places.

\subsection{Predicted Maps}

\textbf{Qualitative examples.} In the upper scene of \cref{fig:navsim_examples}, ego turns left. The
original-split models follow the training drives (pink) closely and pass safely, losing only some
ego progress (EP). The geo split trained models, which never saw this place, turn too far left and
both score zero on drivable area compliance. Interestingly, in their own predicted maps, both models place themselves
off the road (which equals a $DAC=0$ score); LTF even
predicts its front colliding with the pedestrian area.

\textbf{Data leakage on map prediction.} In the upper scenario of \cref{fig:navsim_examples}, the
red circle marks road geometry that the original split trained models predict although no camera
sees it; this information can only come from training drives through the
same location, i.e., from \textit{data leakage} of the training set into the test set. For models that work with historical
states of their own perception, that behavior could be explained; for LTF and DiffusionDrive,
which use only the current frame, only data leakage explains it. The lower scenario in~\cref{fig:navsim_examples} shows similar effects, but more drastically. The left intersection
at the left prediction border is perfectly predicted although not visible. The rather
complicated centerline layout on the upcoming intersection is also predicted correctly,
while the geo split trained models miss it completely although there is no occlusion.
The green circles show another sign of memorization. There, the geo split trained models predict
drivable area and the original-split models do not, matching a ground truth that lacks it. They reproduce
the labels of the place even where the cameras show otherwise.

\Cref{tab:R23} quantifies this. On the geo split, walkway IoU more than halves and
lane centerline IoU falls to about a third. On the 65\,\% of the map that no
camera sees, the original-split models predict the road almost as well as on visible ground (DiffusionDrive $0.65$ vs.\ $0.66$, LTF $0.59$ vs.\ $0.63$), and
the lane centerlines even better. The geo split trained models degrade there (road $0.54 \rightarrow
0.48$ and $0.54 \rightarrow 0.40$). Across all NSIT scenes, the original-split models thus predict
unseen road as if they had seen it, so the examples in \cref{fig:navsim_examples} are no exceptions. Worse maps can cost PDMS when
a plan ends on falsely predicted road, but as the upper example in \cref{fig:navsim_examples} shows,
models can drive well with poor maps, so the PDMS reflects map quality only loosely. %

\subsection{nuScenes}

On nuScenes~\cite{nuscenes}, Lilja \etal~\cite{lilja_geosplit} show that online mapping models lose
more than half of their mAP under a geographic split (MapTRv2~\cite{maptrv2} $-56\,\%$,
StreamMapNet2D~\cite{streammapnet} $-58\,\%$; \cref{tab:spdr}). SparseDrive~\cite{sun_sparsedrive_2025},
which learns mapping as an auxiliary task, drops by $59\,\%$ to the same level. We attribute this
to its map decoder, which builds polylines from initializations learned on the training data, so
that map elements seen in training are memorized well. The mapping drop carries over to planning:
SparseDrive's average collision rate rises from 0.08\,\% to 0.79\,\%, and its average L2 error
from 0.61\,m to 0.76\,m (+25\,\%).

\Cref{fig:curbside} shows a scene in Singapore with a drop-off zone to the left. The place appears
in several logs with different routes: in some, the ego vehicle turns into the drop-off zone, in
others, it proceeds straight. Trained on the original split, SparseDrive predicts the drop-off zone
geometry before the zone is visible (\cref{fig:curbside_straight}); trained on the geo split, it
predicts only a straight road. When entering the zone (\cref{fig:curbside_turn}), both models predict
boundary polylines, but the original-split model's prediction is clearly better, consistent with the
mAP difference.

%% file: sec/tab/tab_eval_schemes.tex
\providecolor{origsplit}{HTML}{C0392B}
\providecolor{geosplit}{HTML}{CC79A7}
\providecolor{seennone}{HTML}{9399b2}

\begin{table}[t]
  \caption{NAVSIMv1 samples by their set in the original (rows) and the geo split (columns).
  Shading: training data of the original model (\legendbox{origsplit!25}), the geo model
  (\legendbox{geosplit!25}), or both (\legendbox{origsplit!50!geosplit!55}). $\ast$/$\dagger$: test
  sets in OVG. Outlined (\legendoutline): NSIT samples. Grey: unseen by both, but in no test set.}\label{tab:evaluation_schemes}
  \centering
  \small
  \setlength{\tabcolsep}{4pt}
  \setlength{\arrayrulewidth}{0.9pt}
  \setlength{\aboverulesep}{0pt}
  \setlength{\belowrulesep}{0pt}
  \renewcommand{\arraystretch}{1.3}
  \begin{tabular}{@{}llccc}
    \toprule
    & & \multicolumn{3}{c}{Geo split} \\
    \cmidrule(l){3-5}
    & & \textcolor{geosplit!75!black}{train\textsubscript{geo}} & val\textsubscript{geo} & test\textsubscript{geo}\,$\dagger$ \\
    \midrule
    \multirow{3}{*}{\rotatebox[origin=c]{90}{Orig.}}
    & \textcolor{origsplit!75!black}{train\textsubscript{orig}}
      & \cellcolor{origsplit!50!geosplit!55}65,796 & \cellcolor{origsplit!25}9,109 & \cellcolor{origsplit!25}9,816 \\
    \cline{5-5}
    & val\textsubscript{orig}
      & \cellcolor{geosplit!25}10,684 & \textcolor{gray}{3,859} & \multicolumn{1}{|c|}{3,418} \\
    \cline{4-4}
    & test\textsubscript{orig}\,$\ast$
      & \cellcolor{geosplit!25}7,450 & \multicolumn{1}{|c}{2,549} & \multicolumn{1}{c|}{1,974} \\
    \cline{4-5}
    \bottomrule
  \end{tabular}
\end{table}

%% file: sec/tab/R02-03.tex
\begin{table}[t]
  \caption{NAVSIMv1 performance, measured in PDMS, for OVG/NSIT evaluation schemes. Orig./Geo: trained on the original/geo split. Differences to the original split trained model in PDMS points in parentheses.}
  \label{tab:R02-03}
  \centering
  \small
  \providecommand{\rel}[1]{{\scriptsize(\textbf{#1})}}
  \setlength{\tabcolsep}{4pt}
  \begin{tabular*}{\linewidth}{@{\extracolsep{\fill}}lcccc@{}}
    \toprule
    Scheme     & \multicolumn{2}{c}{OVG} & \multicolumn{2}{c}{NSIT} \\
    \cmidrule(lr){2-3} \cmidrule(l){4-5}
    Model & Orig. & Geo & Orig. & Geo \\
    \midrule
    DiffusionDrive~\cite{diffusiondrive} & 87.7  & 85.6 \rel{$-$2.1}  & 87.45 & 81.39 \rel{$-$6.1} \\
    LTF~\cite{ltf}            & 80.7  & 80.4 \rel{$-$0.3}  & 81.02 & 77.19 \rel{$-$3.8} \\
    DrivoR~\cite{drivor}         & 93.07 & 91.66 \rel{$-$1.4} & 93.16 & 89.72 \rel{$-$3.4} \\
    Ego-MLP~\cite{egomlp}        & 64.9  & 67.2 \rel{$+$2.3}  & 64.34 & 62.82 \rel{$-$1.5} \\
    \bottomrule
  \end{tabular*}
\end{table}

%% file: sec/tab/R10.tex
\begin{table}[t]
  \caption{Share of NSIT scenes with PDMS $= 0$, and mean PDMS of the remaining scenes.}\label{tab:R10}
  \centering
  \small
  \setlength{\tabcolsep}{4pt}
  \begin{tabular*}{\linewidth}{@{\extracolsep{\fill}}lcccc@{}}
    \toprule
     & \multicolumn{2}{c}{PDMS $= 0$} & \multicolumn{2}{c}{PDMS of other scenes} \\
    \cmidrule(lr){2-3} \cmidrule(l){4-5}
    Model & Orig. & Geo & Orig. & Geo \\
    \midrule
    DiffusionDrive & 5.4\%  & 10.6\% & 92.5 & 91.1 \\
    LTF            & 12.1\% & 15.1\% & 92.1 & 90.9 \\
    DrivoR         & 2.7\%  & 5.5\%  & 95.7 & 95.0 \\
    Ego-MLP        & 28.4\% & 30.1\% & 89.9 & 89.9 \\
    \bottomrule
  \end{tabular*}
\end{table}

%% file: sec/tab/R38.tex
\begin{table}[t]
  \caption{Failure causes of the models measured in PDMS difference:
  $\text{PDMS}_\text{orig} - \text{PDMS}_\text{geo}$. \textit{DDrive =
  DiffusionDrive.}}\label{tab:R38}
  \centering
  \small
  \setlength{\tabcolsep}{2pt}
  \begin{tabular*}{\linewidth}{@{\extracolsep{\fill}}lcccc@{}}
    \toprule
    Cause & DDrive & LTF & DrivoR & Ego-MLP \\
    \midrule
    Road exit $<$0.5\,m       & $+$1.12 & $-$0.03 & $+$1.76 & $+$0.01 \\
    Road exit $\geq$0.5\,m    & $+$2.73 & $+$1.59 & $+$1.17 & $+$1.13 \\
    Collision on road         & $+$1.36 & $+$1.44 & $-$0.19 & $+$0.54 \\
    TTC only                  & $+$0.78 & $+$0.59 & $+$0.28 & $-$0.01 \\
    Progress                  & $+$0.02 & $+$0.24 & $+$0.41 & $-$0.17 \\
    \midrule
    Gap                       & $+$6.01 & $+$3.82 & $+$3.44 & $+$1.52 \\
    \bottomrule
  \end{tabular*}
\end{table}

%% file: sec/tab/R16.tex
\begin{table}[t]
  \caption{Training-drive effect per city on NSIT: how much larger the gap
  $\text{PDMS}_\text{orig} - \text{PDMS}_\text{geo}$ is at places with a training drive than at places
  without one. Bold: the 95\,\% interval excludes zero (intervals in \cref{sec:app_nsit_drive}). $n$: scenes without
  a training drive; Las Vegas (grey) has too few. DDrive: DiffusionDrive.}\label{tab:R16}
  \centering
  \small
  \setlength{\tabcolsep}{3pt}
  \begin{tabular*}{\linewidth}{@{\extracolsep{\fill}}lrcccc@{}}
    \toprule
    City & $n$ & DDrive & LTF & DrivoR & Ego-MLP \\
    \midrule
    Singapore & 739 & $\mathbf{+6.1}$ & $\mathbf{+7.1}$ & $\mathbf{+3.7}$ & $-0.8$ \\
    Boston & 182 & $\mathbf{+5.3}$ & $\mathbf{+7.6}$ & $+1.2$ & $\mathbf{-3.5}$ \\
    Pittsburgh & 210 & $\mathbf{+13.4}$ & $\mathbf{+8.7}$ & $\mathbf{+7.7}$ & $-1.4$ \\
    \textcolor{gray}{Las Vegas} & \textcolor{gray}{41} & \textcolor{gray}{$-2.0$} & \textcolor{gray}{$+2.2$} & \textcolor{gray}{$+10.4$} & \textcolor{gray}{$-1.5$} \\
    \midrule
    All & 1{,}172 & $\mathbf{+6.2}$ & $\mathbf{+7.7}$ & $\mathbf{+3.6}$ & $+0.8$ \\
    \bottomrule
  \end{tabular*}
\end{table}

%% file: sec/tab/R29.tex
\begin{table}[t]
  \caption{NSIT PDMS gap $\text{PDMS}_\text{orig} - \text{PDMS}_\text{geo}$ with and without the map loss,
    seed~0 only. Bold: the 95\,\% interval excludes zero.}\label{tab:R29}
  \centering
  \small
  \begin{tabular*}{\linewidth}{@{\extracolsep{\fill}}lccc@{}}
    \toprule
          & \multicolumn{2}{c}{Map loss} & \\
    \cmidrule(lr){2-3}
    Model & With & Without & Difference \\
    \midrule
    DiffusionDrive~\cite{diffusiondrive} & $+$5.63 & $+$5.48 & $-0.2$ \\
    LTF~\cite{ltf}            & $+$3.45 & $+$5.81 & $\mathbf{+2.4}$ \\
    \bottomrule
  \end{tabular*}
\end{table}

%% file: sec/tab/R23.tex
\begin{table}[t]
  \caption{Map IoU per class on NSIT, on the whole map and split by camera visibility (not visible:
  outside all camera images or occluded in the LiDAR scan; 65\,\% of the map). All drops exclude zero.}\label{tab:R23}
  \centering
  \small
  \begin{tabular}{@{}lcccc@{}}
    \toprule
     & \multicolumn{2}{c}{DiffusionDrive} & \multicolumn{2}{c}{LTF} \\
    \cmidrule(lr){2-3} \cmidrule(l){4-5}
    Class & Orig. & Geo & Orig. & Geo \\
    \midrule
    Road & 0.657 & 0.511 & 0.616 & 0.470 \\
    \quad visible & 0.659 & 0.538 & 0.634 & 0.538 \\
    \quad not visible & 0.654 & 0.480 & 0.594 & 0.396 \\
    \addlinespace
    Walkway & 0.488 & 0.213 & 0.406 & 0.178 \\
    \quad visible & 0.509 & 0.248 & 0.438 & 0.217 \\
    \quad not visible & 0.473 & 0.189 & 0.383 & 0.152 \\
    \addlinespace
    Lane centre line & 0.337 & 0.116 & 0.253 & 0.092 \\
    \quad visible & 0.324 & 0.129 & 0.245 & 0.106 \\
    \quad not visible & 0.354 & 0.097 & 0.263 & 0.073 \\
    \bottomrule
  \end{tabular}
\end{table}

%% file: sec/tab/spdr.tex
\begin{table*}[t]
  \caption{SparseDrive~\cite{sun_sparsedrive_2025} on nuScenes with the original split and the geo split
    (+ Geo) of Lilja \etal~\cite{lilja_geosplit}. Mapping AP in \%, prediction for cars. Map-only
    baselines from~\cite{lilja_geosplit}: MapTRv2~\cite{maptrv2} (ResNet-50, LSS lifting, DETR decoder) and
    StreamMapNet2D~\cite{streammapnet} (BEVFormer lifting), on the validation set of the geo split.}\label{tab:spdr}
  \centering
  \footnotesize
  \setlength{\tabcolsep}{2pt}
  \begin{tabular*}{\textwidth}{@{\extracolsep{\fill}}l cc ccc cccc cc cc cc@{}}
    \toprule
    & \multicolumn{2}{c}{Detection} & \multicolumn{3}{c}{Tracking} & \multicolumn{4}{c}{Mapping AP$\uparrow$}
      & \multicolumn{2}{c}{Prediction$\downarrow$} & \multicolumn{2}{c}{Plan. L2$\downarrow$} & \multicolumn{2}{c}{Plan. CR$\downarrow$} \\
    \cmidrule(lr){2-3} \cmidrule(lr){4-6} \cmidrule(lr){7-10} \cmidrule(lr){11-12} \cmidrule(lr){13-14} \cmidrule(l){15-16}
     & mAP$\uparrow$ & NDS$\uparrow$ & AMOTA$\uparrow$ & AMOTP$\downarrow$ & Recall$\uparrow$
      & Ped. & Div. & Bdry. & Mean
      & minADE & minFDE
      & 3s & Avg. & 3s & Avg. \\
    \midrule
    MapTRv2~\cite{maptrv2}              & -- & -- & -- & -- & -- & 59.1 & 61.8 & 63.7 & 61.5 & -- & -- & -- & -- & -- & -- \\ %
    MapTRv2~\cite{maptrv2} + Geo        & -- & -- & -- & -- & -- & 26.5 & 20.9 & 32.6 & 26.7 & -- & -- & -- & -- & -- & -- \\ %
    StreamMapNet2D~\cite{streammapnet}       & -- & -- & -- & -- & -- & 61.0 & 64.5 & 62.3 & 62.6 & -- & -- & -- & -- & -- & -- \\ %
    StreamMapNet2D~\cite{streammapnet} + Geo & -- & -- & -- & -- & -- & 25.8 & 23.0 & 29.5 & 26.1 & -- & -- & -- & -- & -- & -- \\ %
    \midrule
    SparseDrive~\cite{sun_sparsedrive_2025} & 0.418 & 0.525 & 0.386 & 1.254 & 0.499 & 49.9 & 57.0 & 58.4 & 55.1 & 0.620 & 0.990 & 0.960 & 0.610 & 0.180 & 0.080 \\
    SparseDrive + Geo                       & 0.328 & 0.444 & 0.277 & 1.445 & 0.392 & 19.5 & 19.9 & 27.4 & 22.3 & 0.756 & 1.224 & 1.214 & 0.762 & 1.474 & 0.786 \\
    \bottomrule
  \end{tabular*}
\end{table*}

%% file: sec/5_conclusion.tex
\FloatBarrier %
\section{Conclusion}

We propose a geo split for NAVSIMv1 that makes the training, validation, and test sets
geographically disjoint, and compare models trained on either split on scenes neither has seen in
training. The geographic overlap of the original split inflates the PDMS of all sensor-based models
we test, including DrivoR, which has no explicit map representation. The advantage of the original
split trained models comes from places where a training drive passed; at all other places, they are
no better. At visited places, the geo split trained models more often leave the
road, collide, or come close to other objects, and more scenes score zero. Models trained on the
original split also predict map elements they cannot see, on NAVSIMv1 and on nuScenes; on NAVSIMv1,
almost as well as visible ones; on nuScenes, SparseDrive's planning degrades along with its maps.
The gap persists when the auxiliary map loss is removed, so the models memorize locations even
without being supervised to. This raises the question whether and how end-to-end models should be
supervised on maps, and how much their planning relies on map perception.
Assessing the planning capability of a model on new places requires a split that is geographically
disjoint, which through our proposal is now possible for NAVSIMv1. We recommend reporting NAVSIMv1 results on
our geo split alongside navtest, so that improvements can be separated from memorized locations.
Our evaluation covers logged
scenes; whether location memorization also affects scores on re-rendered scenes, as in NAVSIMv2,
remains open.

%% file: sec/6_ai_statement.tex
\section*{LLM Usage Statement}
We used Claude (Anthropic) for editing the text, preparing tables and figures,
and checking statements against cited sources. All analyses, results, and claims were verified
by the authors, who take full responsibility for the content.

%% file: sec/X_suppl.tex
\onecolumn %
{\centering\Large\textbf{Don't Test Where You Trained: Geographic Overlap Inflates End-to-End Driving Scores}\\\vspace{0.5em}Supplementary Material\\\vspace{1.0em}\par}
\appendix
\crefalias{section}{appendix}

\counterwithin{table}{section}\renewcommand{\thetable}{\thesection\arabic{table}}
\counterwithin{figure}{section}\renewcommand{\thefigure}{\thesection\arabic{figure}}

\section{Original maps with data overlaps and proposed geo split maps}\label{sec:app_maps}

\begin{figure*}[!htbp]
  \centering
  \includegraphics[width=1.0\textwidth]{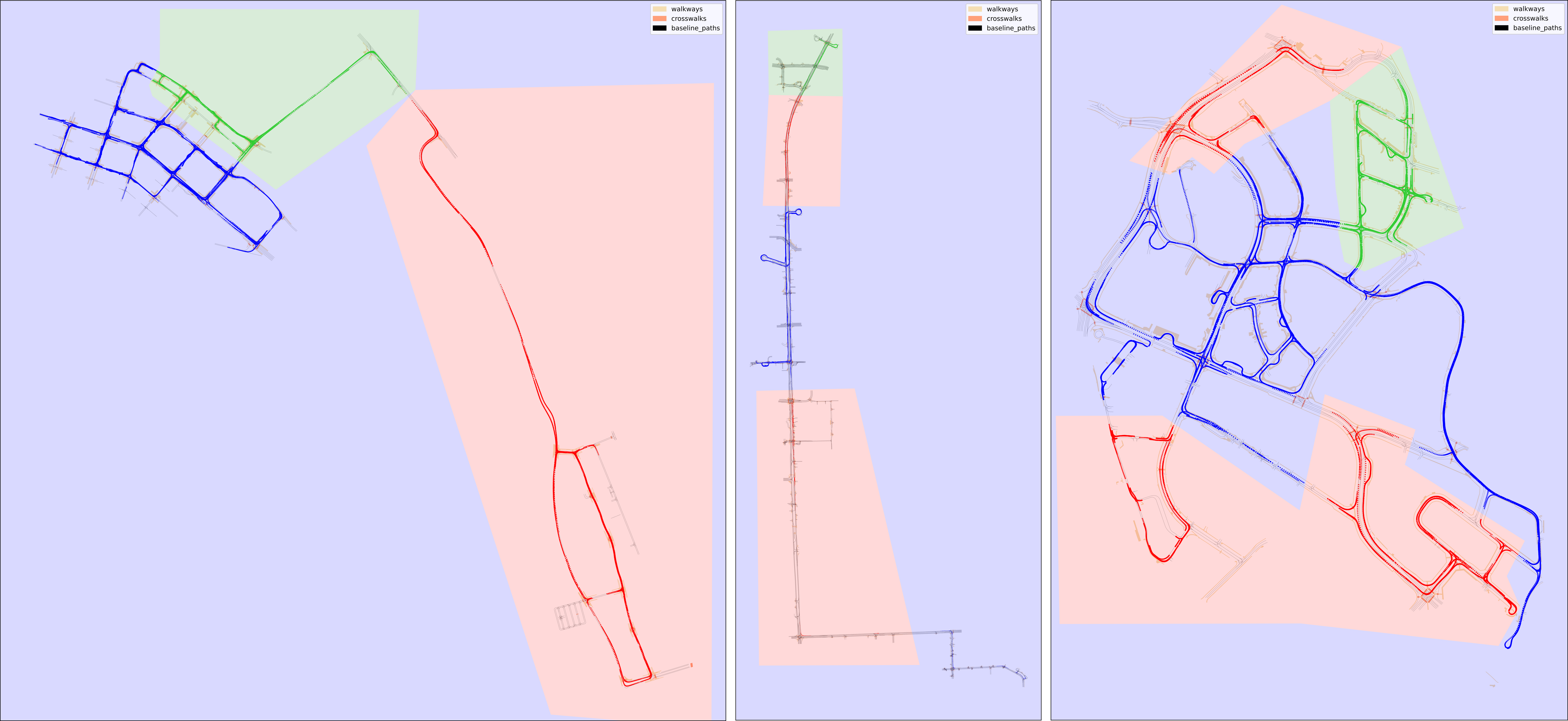}
  \caption{Left to right: Pittsburgh-Hazelwood, Las Vegas, and Singapore maps with the proposed geo split. Training samples are colored blue (\legendline{mplblue}{1.0}), validation samples in red (\legendline{mplred}{1.0}), and test samples in green (\legendline{mplgreen}{1.0}).}\label{fig:app_other_maps}
\end{figure*}

Figure~\ref{fig:app_other_maps} shows the other three cities of our proposed NAVSIMv1 geo
split. In the maps, the \texttt{validation} subset lies in the areas of red polygons (\legendbox{mplred!15}),
the \texttt{test} subset in the green polygons area (\legendbox{mplgreen!15}), and the remaining
background, colored in blue (\legendbox{mplblue!15}), forms the \texttt{training} set. As for Boston
(\cref{fig:boston_geo_map}), whole regions are assigned to one set. For every subset,
we distribute areas with intersections and other challenging situations to all splits, especially for
Singapore, which is the only location with left-hand traffic. We colorize the samples of the subsets
in the respective color with equal thickness, blue for training (\legendline{mplblue}{1.0}), red for
validation (\legendline{mplred}{1.0}), and green for test (\legendline{mplgreen}{1.0}) sample data.

\begin{figure*}[!htbp]
  \centering
  \includegraphics[width=0.98\textwidth]{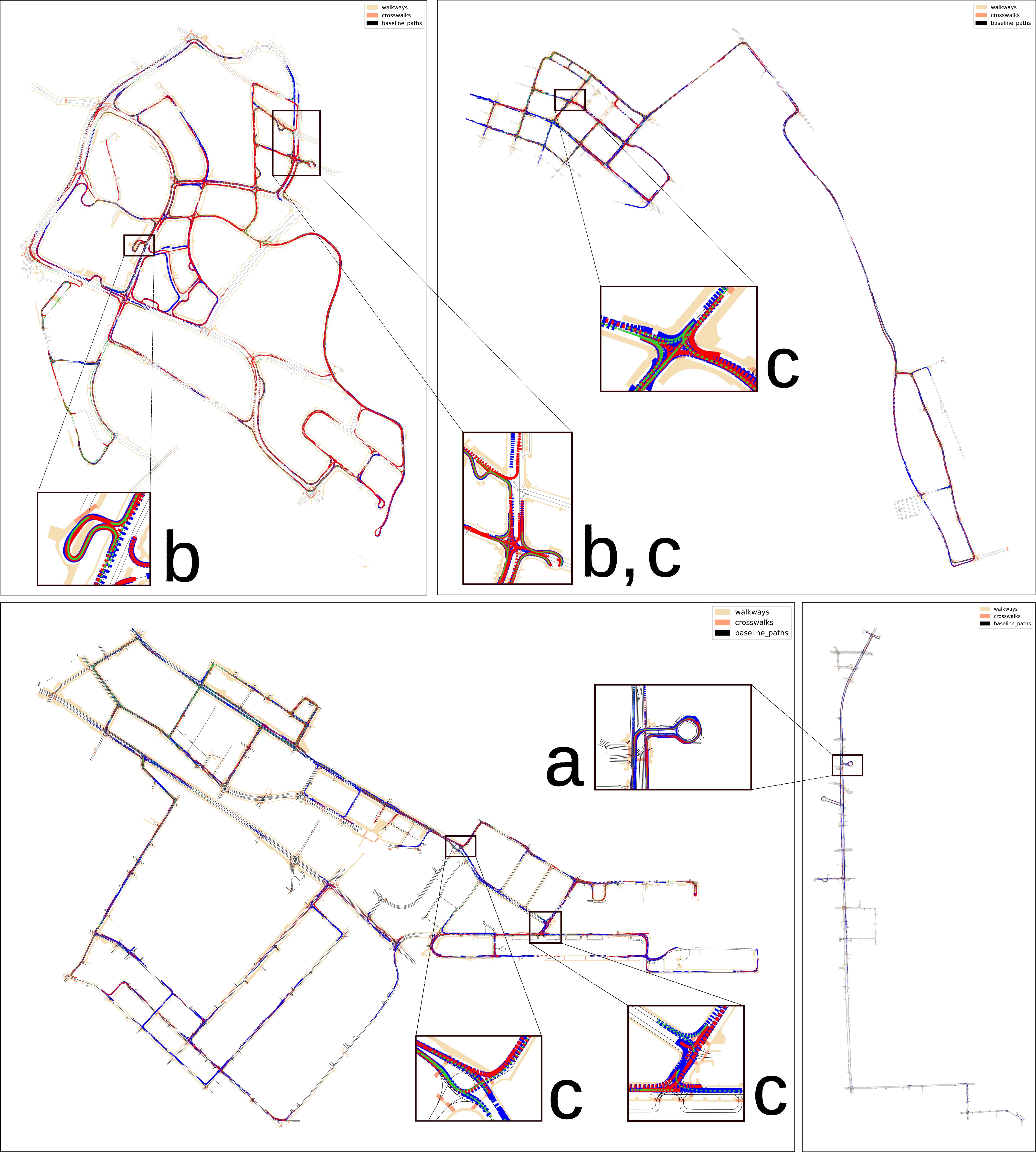}
  \caption{From left to right: Upper row: Singapore, Pittsburgh-Hazelwood. Lower row: Boston, Las Vegas. Data from the
  original split of NAVSIM~\cite{navsim}. Blue lines (\legendline{mplblue}{1.6}) indicate
  samples from the training set, red ones (\legendline{mplred}{1.2})
  from the validation set, green ones
  (\legendline{mplgreen}{0.9}) from the test set. Detail a: Roundabouts, detail b: drop-off zones, detail c: complex intersections.}\label{fig:app_overlap_maps}
\end{figure*}

In~\cref{fig:app_overlap_maps}, we present all samples in the original split~\cite{navsim} distribution visualized
in an overlay fashion. We choose three line widths and colors to highlight the overlap for the subsets. Training data is visualized with blue dashes (\legendline{mplblue}{1.6}),
validation data with red (\legendline{mplred}{1.2}), and test samples with green (\legendline{mplgreen}{0.9}).
Our measured overlap of $89.6\%$ for test samples to training samples within a 4m radius and $80.4\%$ for test to validation
samples is clearly visible when zooming in to arbitrary areas of the maps. Compared to the training set,
the validation set is not independent either, sharing a $83.8\%$ overlap with training samples.
We cut out a selection of challenging scenarios that show a heavy overlap over all three
subsets, e.g., at roundabouts (\cref{fig:app_overlap_maps}, detail a), at drop-off zones (\cref{fig:app_overlap_maps}, detail b), or at multi-lane and complex intersections
(\cref{fig:app_overlap_maps}, detail c).

\FloatBarrier
\section{NSIT PDMS with and without training drive}\label{sec:app_nsit_drive}
\begin{figure*}[!htbp]
  \centering
  \includegraphics[width=0.62\textwidth]{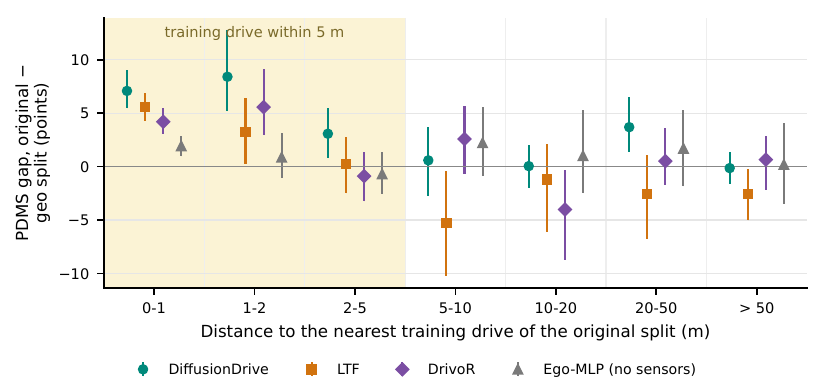}
  \caption{PDMS gap on NSIT by distance to the nearest training drive of the original split, with 95\,\%
  intervals. The gap of the sensor-based models is largest within 2\,m and consistent with zero beyond
  5\,m (shaded: our 5\,m threshold); Ego-MLP shows no trend.}\label{fig:app_gap_distance}
\end{figure*}

\begin{figure*}[!htbp]
  \centering
  \includegraphics[width=0.62\textwidth]{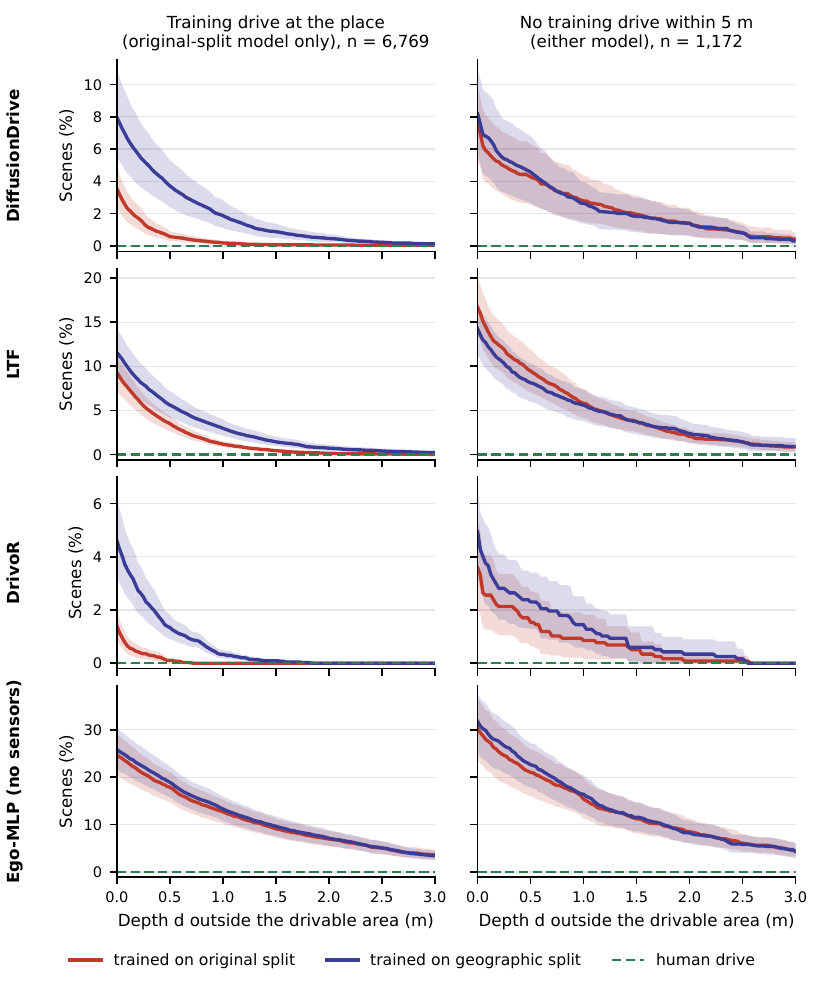}
  \caption{Road exits on NSIT scenes with and without an original-split training drive at the place, absolute shares per split (counterpart to \cref{fig:nsit_combined}a).}\label{fig:app_exit_depth_abs}
\end{figure*}

\begin{figure*}[!htbp]
  \centering
  \includegraphics[width=0.62\textwidth]{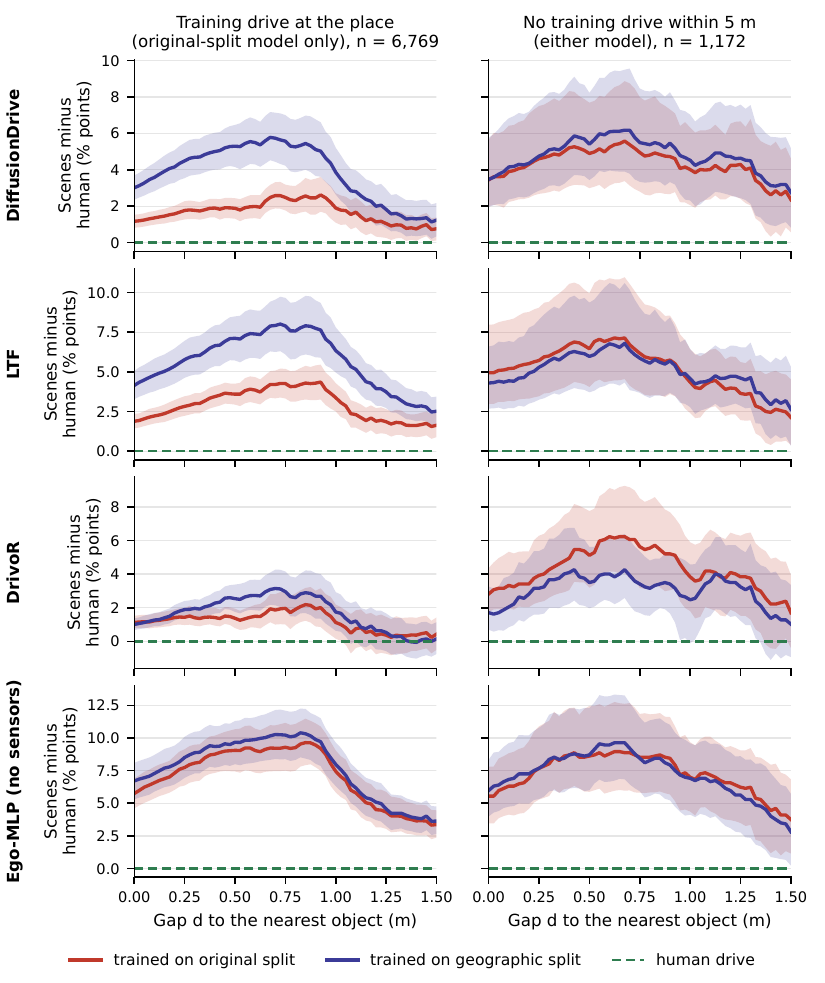}
  \caption{Closest approach on NSIT scenes with and without an original-split training drive at the place, share minus the human drive (counterpart to \cref{fig:nsit_combined}b).}\label{fig:app_closest_abs}
\end{figure*}

An NSIT scene has a training drive if a training scene of the original split lies within 5\,m of it
with a heading difference of at most $15^\circ$ (6,769 scenes); the other 1,172 scenes have none.
\Cref{tab:R16_full} gives the gap per city with 95\,\% intervals; Las Vegas has only 41 scenes without
a training drive, so its row is not informative. LTF's gap without a training drive is negative only
because of Singapore. \Cref{fig:app_gap_distance} breaks the gap down by the distance to the nearest training
drive. It is largest within 2\,m and consistent with zero beyond 5\,m, which supports the 5\,m
threshold. \Cref{fig:app_exit_depth_abs,fig:app_closest_abs} show the two splits separately instead of their
difference (\cref{fig:nsit_combined}).

\input{sec/tab/R16_full} %

\FloatBarrier
\section{Seed robustness}\label{sec:app_seeds}
DiffusionDrive and LTF are trained with three seeds on each split; \cref{tab:seed_robustness} pairs
every original-split seed with every geo-split seed. DrivoR and Ego-MLP are single runs per split. In all 18 pairings, the NSIT gap and the gap at
places with a training drive are positive, and the gap with a training drive exceeds the gap
without one. The negative gap of LTF without a training drive holds in all 9 pairings.

\input{sec/tab/tab_seed_robustness}

\FloatBarrier
\section{PDMS distribution}\label{sec:app_pdms_hist}
\begin{figure*}[!htbp]
  \centering
  \includegraphics[width=0.62\textwidth]{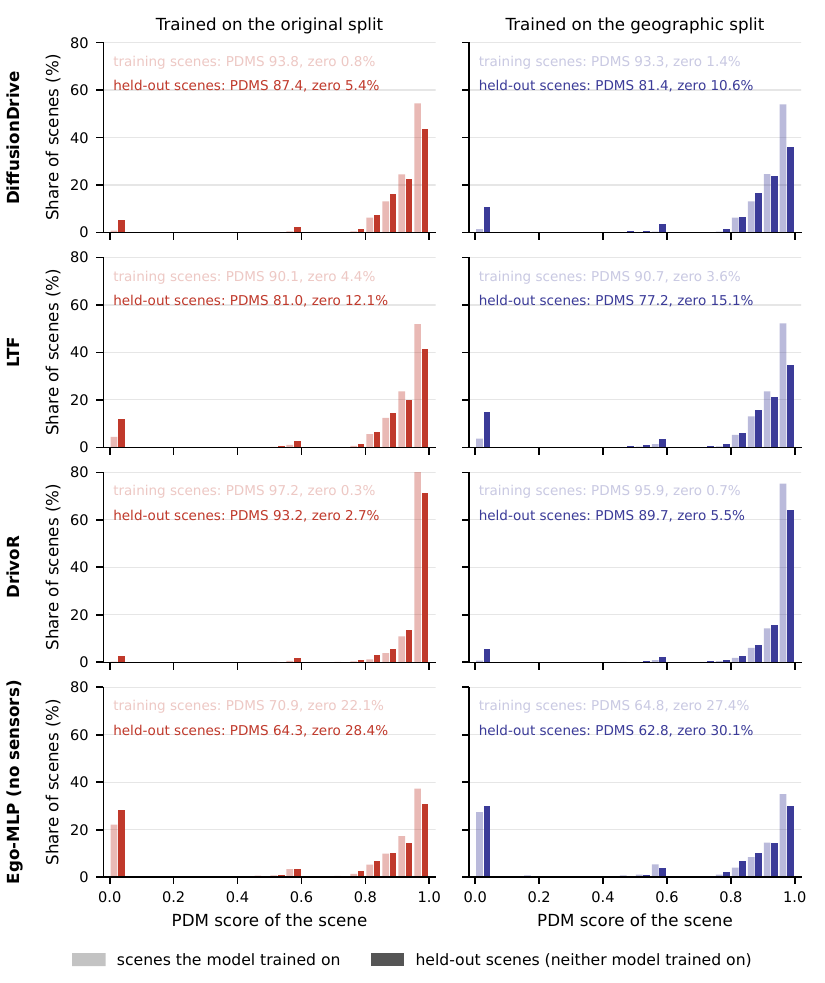}
  \caption{PDMS distribution per model and training split, on the scenes the model was trained on and on the held-out NSIT scenes (cf.~\cref{tab:R10}).}\label{fig:app_pdms_hist}
\end{figure*}

On the scenes each model was trained on, models trained on either split reach almost the same PDMS (DiffusionDrive
$93.8$ vs.\ $93.3$) with few zero scores ($0.8\,\%$ vs.\ $1.4\,\%$, \cref{fig:app_pdms_hist}). The sensor-based
models therefore fit their training data about equally well on both splits. On NSIT, the share of zero scores grows on the geo split
for the sensor-based models (DiffusionDrive $5.4\,\% \rightarrow 10.6\,\%$, LTF $12.1\,\% \rightarrow 15.1\,\%$, DrivoR $2.7\,\% \rightarrow
5.5\,\%$, \cref{tab:R10}). Ego-MLP has many zero scores on every scene set ($22$ to $30\,\%$).

\FloatBarrier
\section{Other scene sets}\label{sec:app_scene_sets}
\begin{figure*}[!htbp]
  \centering
  \includegraphics[width=0.62\textwidth]{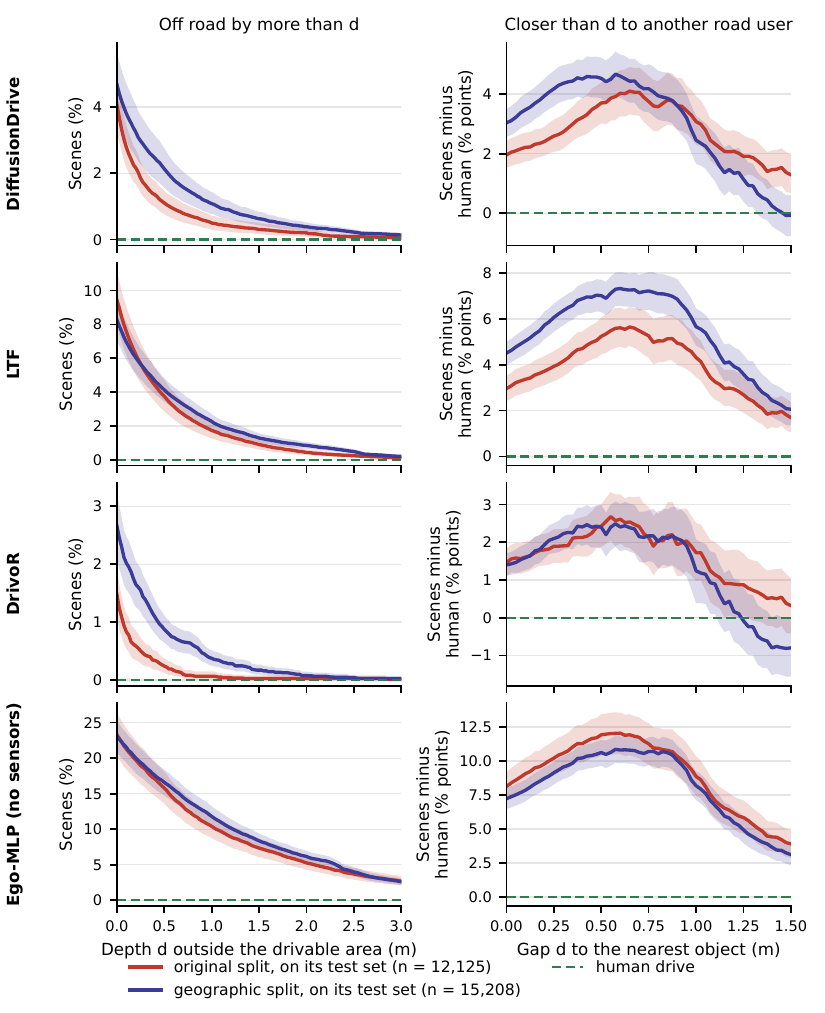}
  \caption{Road exits and closest approach, each model on its own test set (OVG).}\label{fig:app_curves_own}
\end{figure*}

In OVG, each model is evaluated on its own test set (\cref{fig:app_curves_own}). Trained on the geo split,
DiffusionDrive and DrivoR leave the road more often, and DiffusionDrive and LTF come close to other
objects more often. The two test sets differ in composition (\cref{tab:city_composition}),
so this comparison is only indicative.

\begin{figure*}[!htbp]
  \centering
  \includegraphics[width=0.62\textwidth]{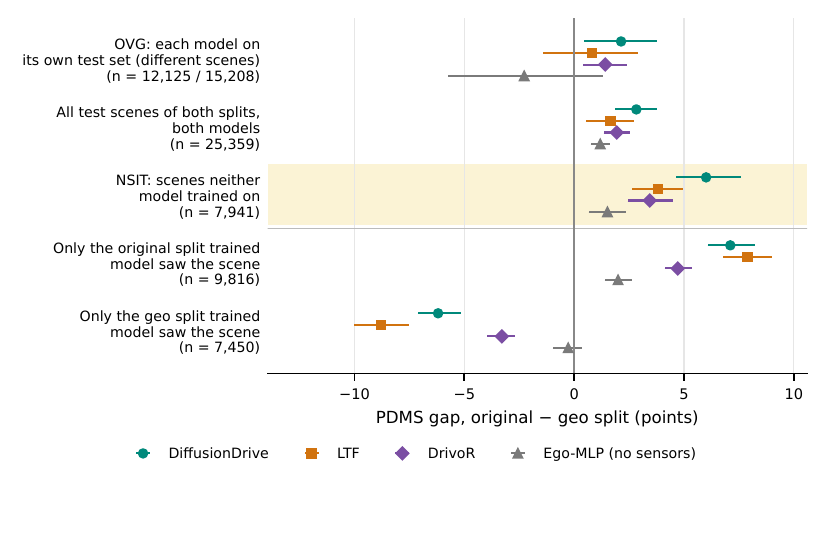}
  \caption{PDMS gap $\text{PDMS}_\text{orig} - \text{PDMS}_\text{geo}$ by scene set, with 95\,\% intervals.
  On scenes that only the original split trained model saw in training, it gains about as much as on NSIT
  scenes at places with a training drive (\cref{tab:R16_full}); on scenes only the geo split trained
  model saw, the gap reverses.}\label{fig:app_gap_scene_sets}
\end{figure*}

\Cref{fig:app_gap_scene_sets} splits the PDMS gap by scene set: OVG, NSIT, and the scenes that only one
of the two models was trained on. On scenes that only the original split trained DiffusionDrive saw in
training, the gap is about 7 points,
about as much as on NSIT scenes at places with a training drive ($+6.9$, \cref{tab:R16_full}). Having seen
the place is worth almost as much as having seen the scene. On scenes only the geo split trained model
saw, the gap reverses.

\FloatBarrier
\section{Map-loss ablation}\label{sec:app_mapablation}
\Cref{tab:map_ablation} lists all runs of the map-loss ablation. For DiffusionDrive, removing the map
loss changes neither model significantly. For LTF, both models lose, the geo split trained one more
($-4.3$ vs.\ $-2.0$ PDMS points), which widens the gap between the splits.

\input{sec/tab/tab_map_ablation}

\FloatBarrier
\section{Validation-set check}\label{sec:app_valval}
NSIT also contains validation scenes of one split, although every NSIT scene lies in the test set of
the other (\cref{tab:evaluation_schemes}). As an additional check, we evaluate the 3,859 scenes that lie in the validation
set of both splits, which no model was trained or selected on, for
DiffusionDrive, LTF, and Ego-MLP (DrivoR was not evaluated on this set). The gap repeats
(\cref{tab:valval}) and is again concentrated at places with a training drive. On this set, the
plain difference between scenes with and without a training drive is confounded by city: almost
all Las Vegas and Pittsburgh scenes have a training drive, and Ego-MLP's gap is large in exactly
these cities. Weighted per city, the difference vanishes for Ego-MLP and remains large for
DiffusionDrive and LTF.

\input{sec/tab/tab_valval}

\FloatBarrier
\section{Pittsburgh held out}\label{sec:app_pittsburgh}
Following Löwens et al.~\cite[Sec.~4.5]{loewens2026memory}, we train DiffusionDrive, LTF, and Ego-MLP
on the original split with every Pittsburgh scene removed (77,269 of 85,109 training scenes), with the
same recipe, epochs, and seeds, and compare them with the original-split models used throughout the
paper (\cref{tab:pittsburgh_holdout}). On navtest, the models trained with Pittsburgh score about
11 PDMS points higher on Pittsburgh scenes for DiffusionDrive and LTF, and all 9 seed pairings are
positive (DiffusionDrive $+9.9$ to $+12.3$, LTF $+5.4$ to $+14.8$). The sensor-free Ego-MLP gains $+3.3$,
and in the other cities the difference stays below one point. On the recorded first-stage scenes of
navhard (NAVSIMv2), the Pittsburgh gap is $+19.4$ and $+16.0$ EPDMS. This is the only result in the paper on
NAVSIMv2. Unlike the two-stage score of \cite[Tab.~7a]{loewens2026memory}, it excludes the second
stage of re-rendered scenes off the logged path and uses non-reactive traffic; the re-rendered
scenes are camera-only, so DiffusionDrive cannot be evaluated on them, and we did not run this stage for the other models.

\input{sec/tab/tab_pittsburgh_holdout}

\FloatBarrier
\section{Qualitative examples}\label{sec:app_qualitative}
Each figure shows a NAVSIMv1 test scene in the layout of \cref{fig:navsim_examples}: (a) ground truth
with the training drives of the original split near the scene (pink), (b, e) ground truth with the plans
of LTF and DiffusionDrive, (c, d, f, g) the BEV maps predicted by the original split and the geo split
trained models, with PDMS and subscores. For the pink lines, every training scene of the original split
within 60\,m of the test scene (excluding the test scene's own drive) is placed at its recorded
position; the positions of each drive are joined in time order and shown in the frame of the test
vehicle.

\subsection*{Following the training drives}

\begin{figure*}[!htbp]
  \centering
  \includegraphics[width=\linewidth]{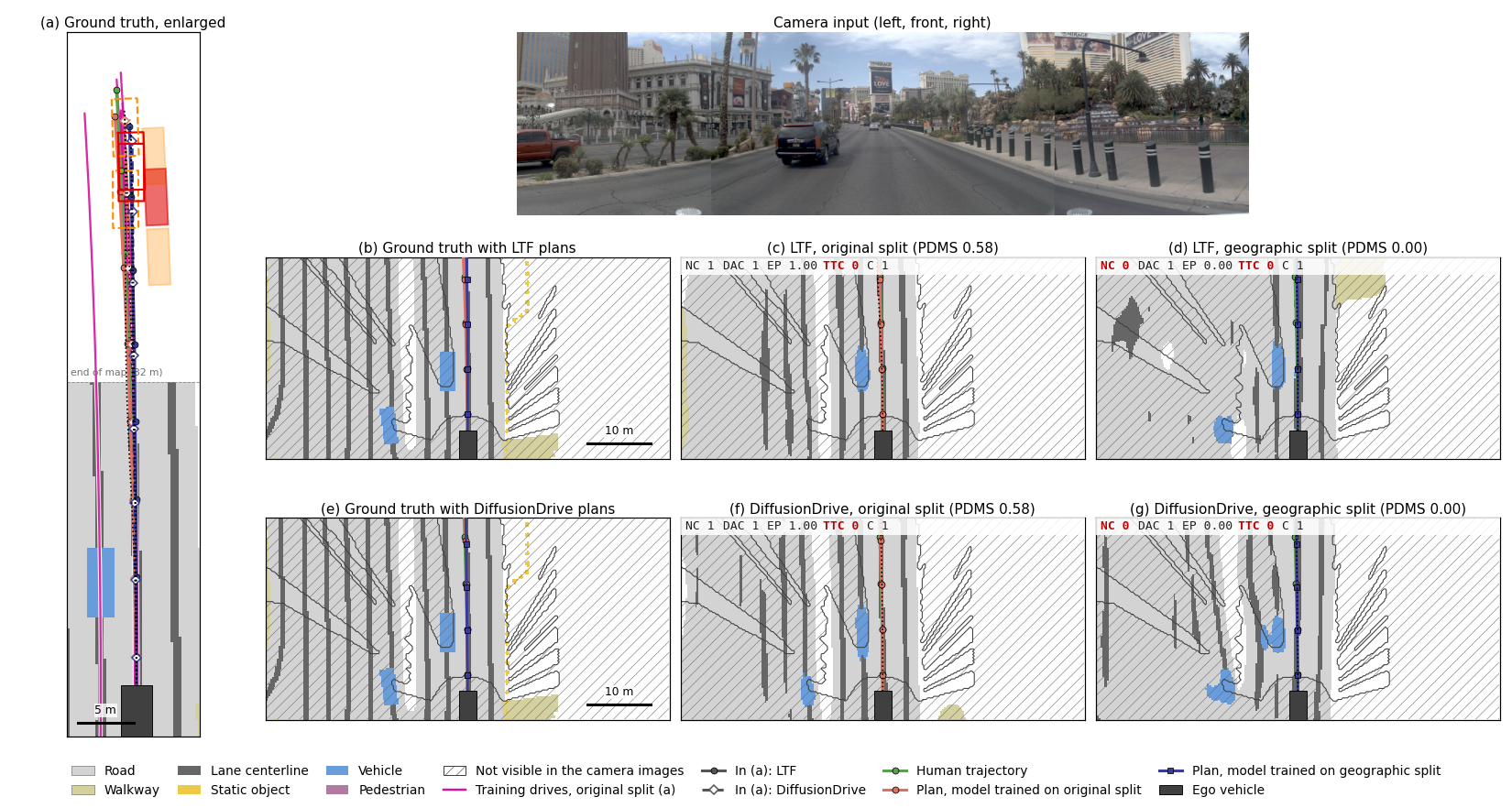}
  \caption{Las Vegas. Beyond the map range, the training drives of the original split follow a slight bend
  (a, pink), which neither the lane centerlines (b, e) nor the camera images indicate. The original split
  trained models follow this bend (c, f; PDMS 0.58). The geo split trained models continue straight and hit
  a vehicle beyond the map range (a, red box; d, g; PDMS 0.00), although their maps are similar.}\label{fig:app_qual_bend_collision}
\end{figure*}

\begin{figure*}[!htbp]
  \centering
  \includegraphics[width=\linewidth]{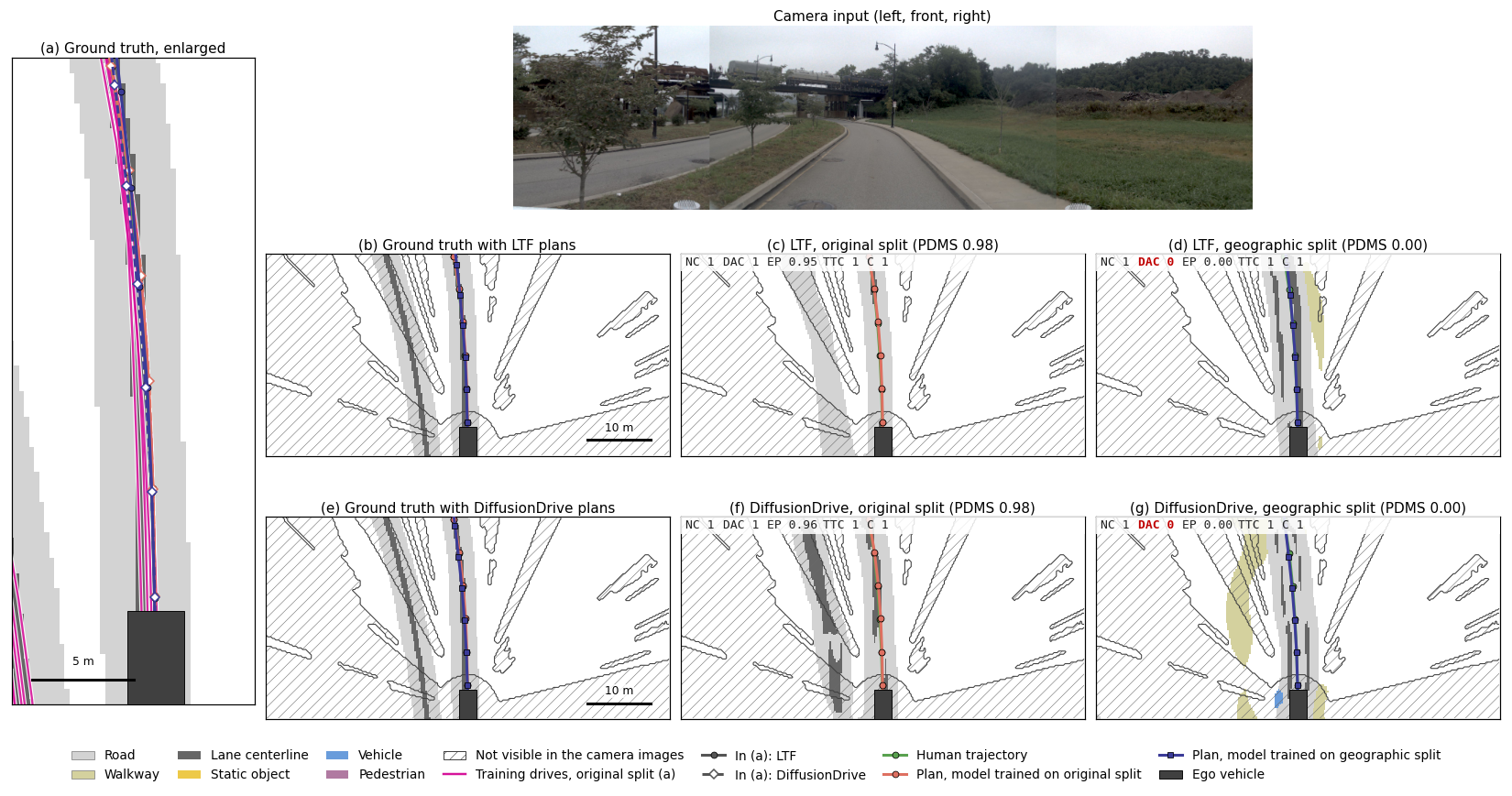}
  \caption{Pittsburgh, narrow road with a slight left bend. The training drives of the original split follow the lane
  closely (a, pink), and so do both original split trained models (c, f; PDMS 0.98). Trained on the geo
  split, both models plan a few decimeters farther left and leave the drivable area (d, g; PDMS 0.00).
  A small difference in the plan yields the largest possible difference in the score.}\label{fig:app_qual_bend_exit}
\end{figure*}

\begin{figure*}[!htbp]
  \centering
  \includegraphics[width=\linewidth]{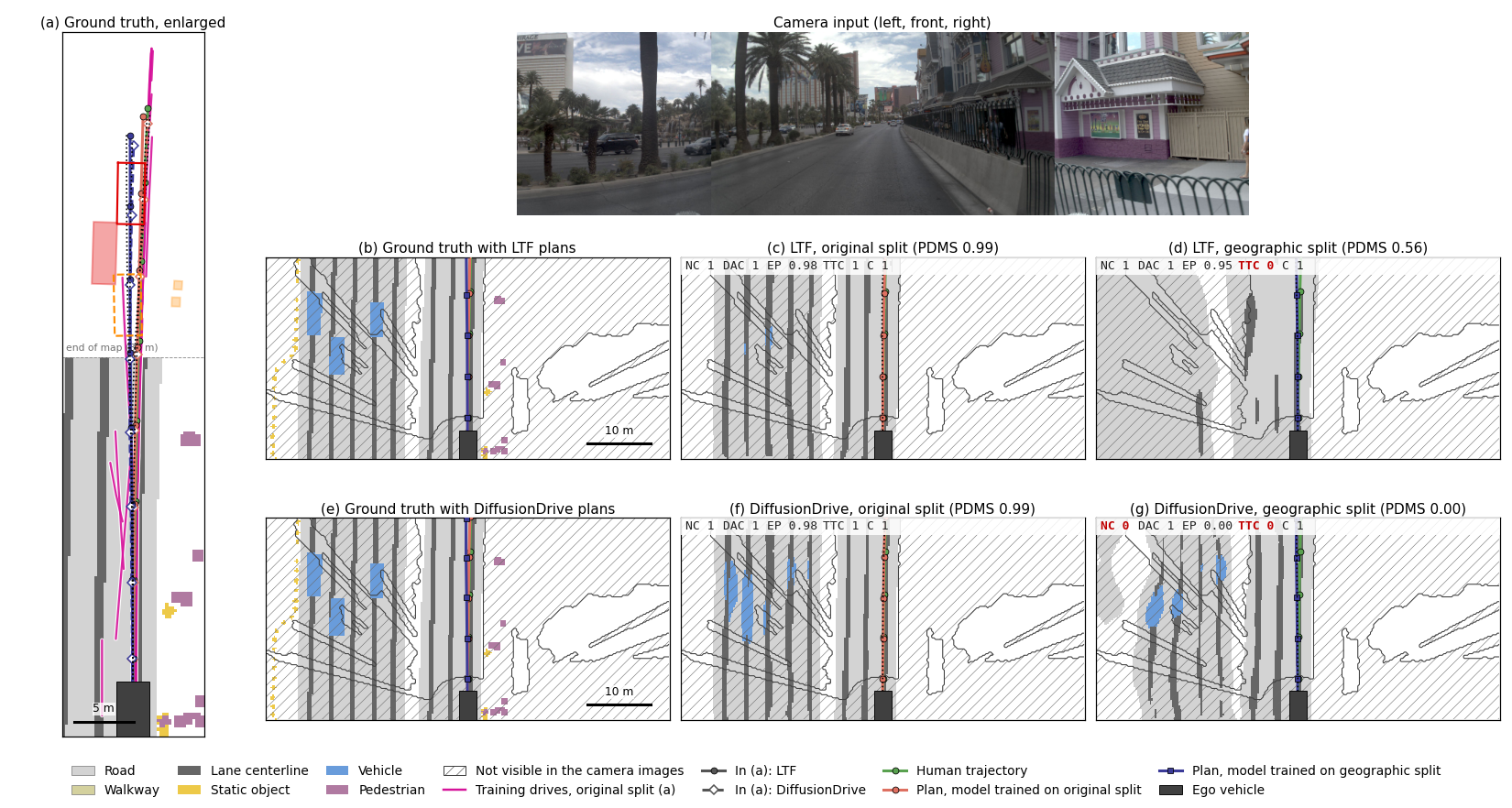}
  \caption{Las Vegas, straight multi-lane road passed by many training drives of the original split (a, pink).
  Trained on the original split, both models predict the lane layout (c, f) and plan like the human driver
  (PDMS 0.99). Trained on the geo split, they predict only a coarse road (d, g) and plan farther ahead;
  DiffusionDrive hits the vehicle in front beyond the map range (a, red box; g; PDMS 0.00).}\label{fig:app_qual_far_collision}
\end{figure*}

\FloatBarrier
\subsection*{Remembered routes and rare maneuvers}

\begin{figure*}[!htbp]
  \centering
  \includegraphics[width=\linewidth]{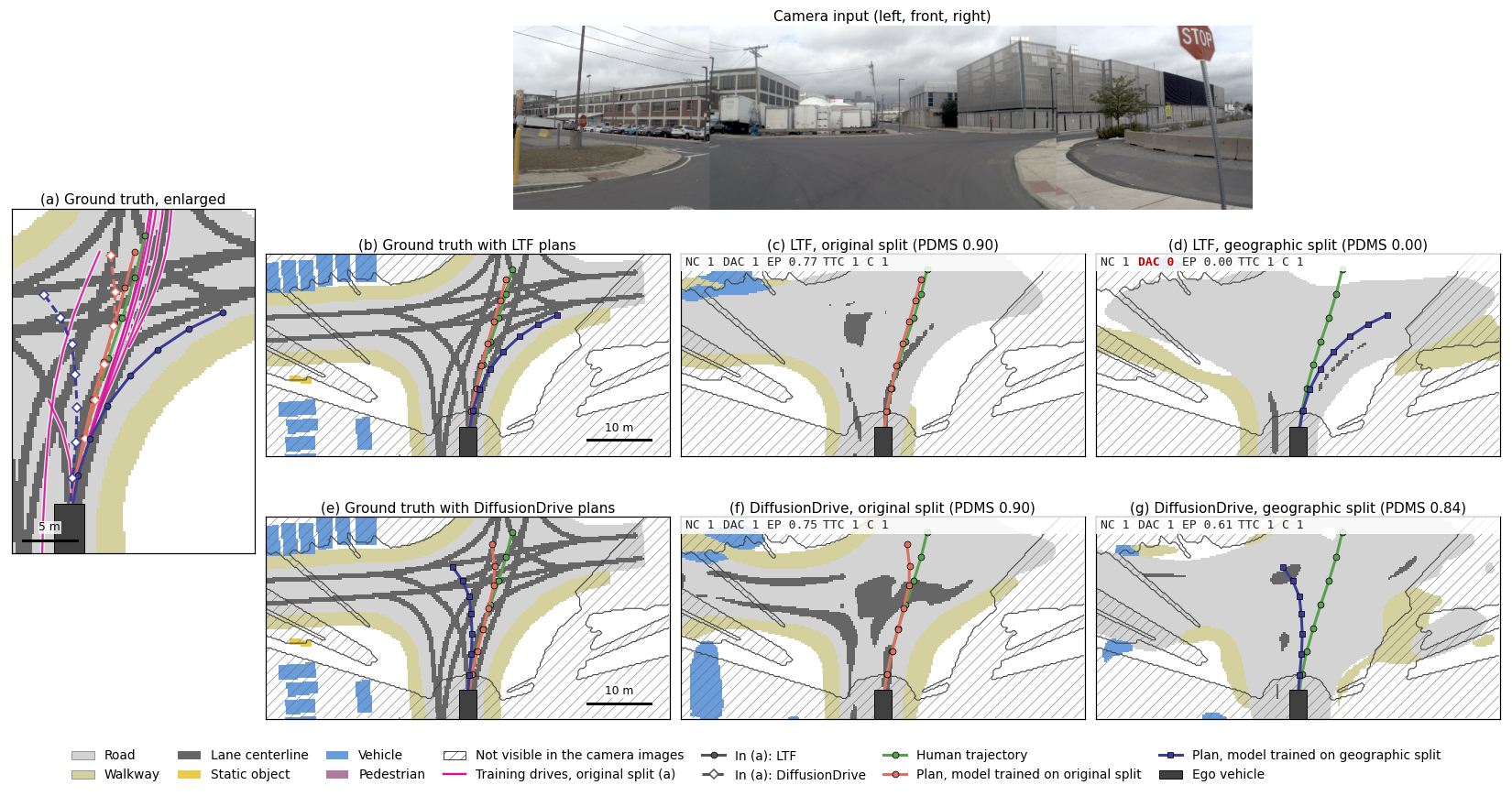}
  \caption{Boston, crossing a wide intersection without lane markings. The training drives of the original split
  take one path across it (a, pink). Trained on the original split, both models follow this path (c, f;
  PDMS 0.90), although the camera images show no markings. Trained on the geo split, LTF picks a different
  mode and turns right, leaving the drivable area (d; PDMS 0.00); DiffusionDrive drifts left (g; PDMS
  0.84). Neither original split trained model considers another route.}\label{fig:app_qual_route_mode}
\end{figure*}

\begin{figure*}[!htbp]
  \centering
  \includegraphics[width=\linewidth]{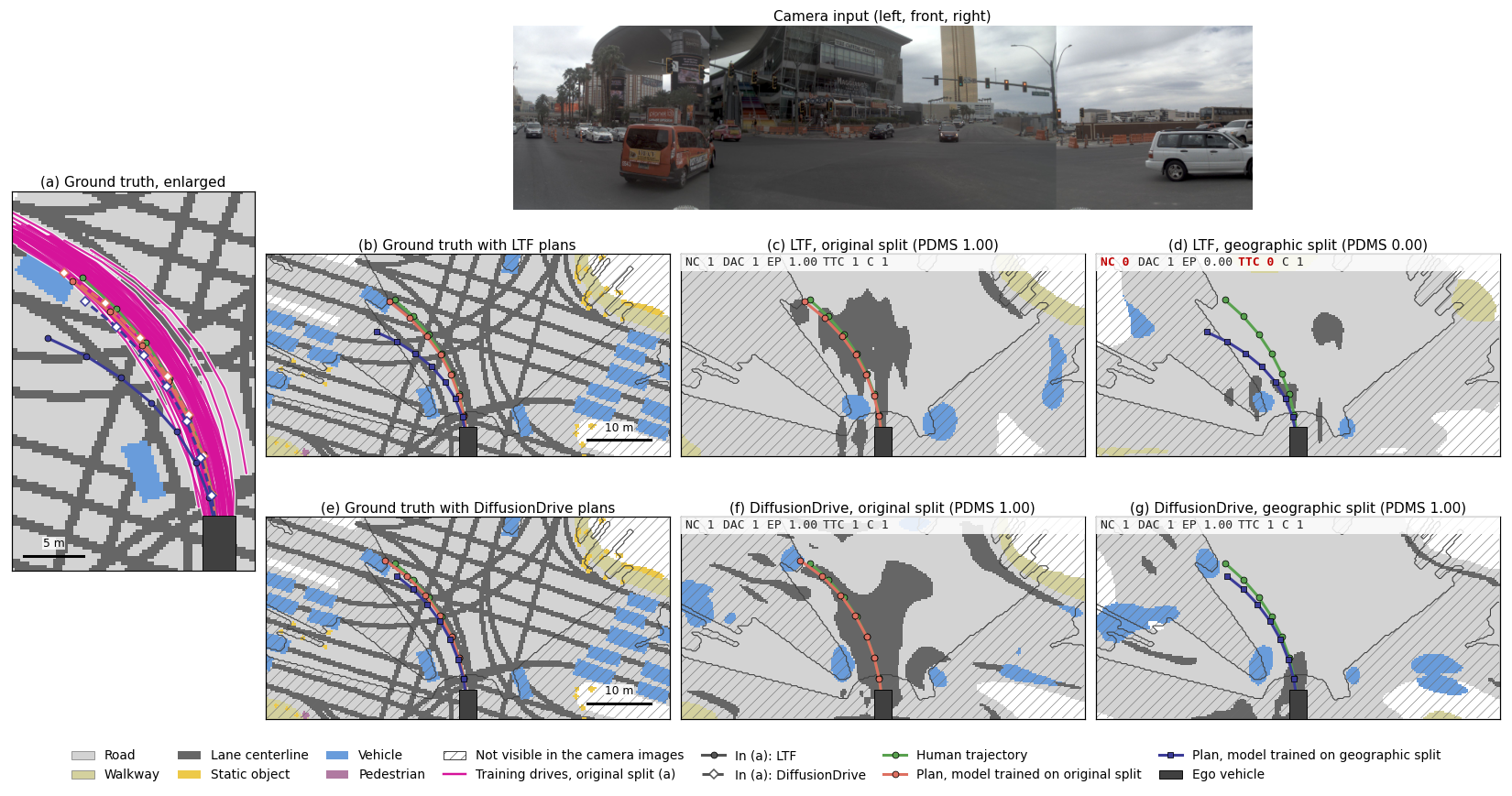}
  \caption{Las Vegas, left turn across a large intersection with many possible lanes (b, e). Training drives of
  the original split pass here densely (a, pink). The original split trained models predict only the one
  turning centerline that these drives use (c, f) and follow it (PDMS 1.00). Trained on the geo split, LTF
  turns too sharply and hits a vehicle (a, d; PDMS 0.00); DiffusionDrive completes the turn (g; PDMS
  1.00).}\label{fig:app_qual_route_centerline}
\end{figure*}

\begin{figure*}[!htbp]
  \centering
  \includegraphics[width=\linewidth]{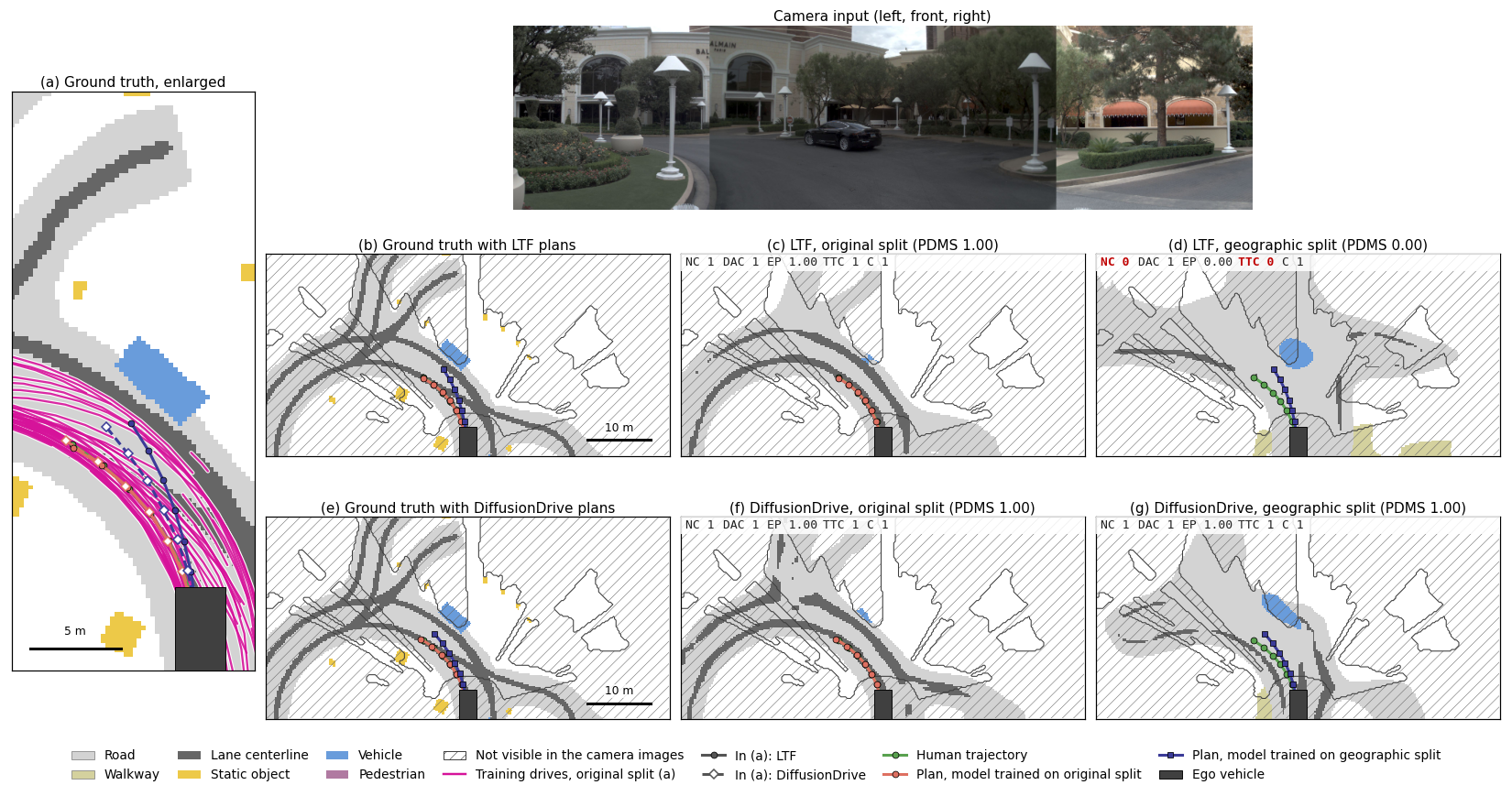}
  \caption{Las Vegas, hotel driveway with a tight left curve, a maneuver rare in the training data. Only the
  original split trained models bend their plans enough to follow the lane (a, b, e; PDMS 1.00), as the
  training drives at this place do (a, pink); they also predict the curve outside the camera view (c, f).
  Both geo split trained models turn too little. LTF runs into the parked car ahead (d; PDMS 0.00), and
  DiffusionDrive cuts the curve without hitting anything (g; PDMS 1.00).}\label{fig:app_qual_tight_curve}
\end{figure*}

\FloatBarrier
\subsection*{Failures independent of the split}

\begin{figure*}[!htbp]
  \centering
  \includegraphics[width=\linewidth]{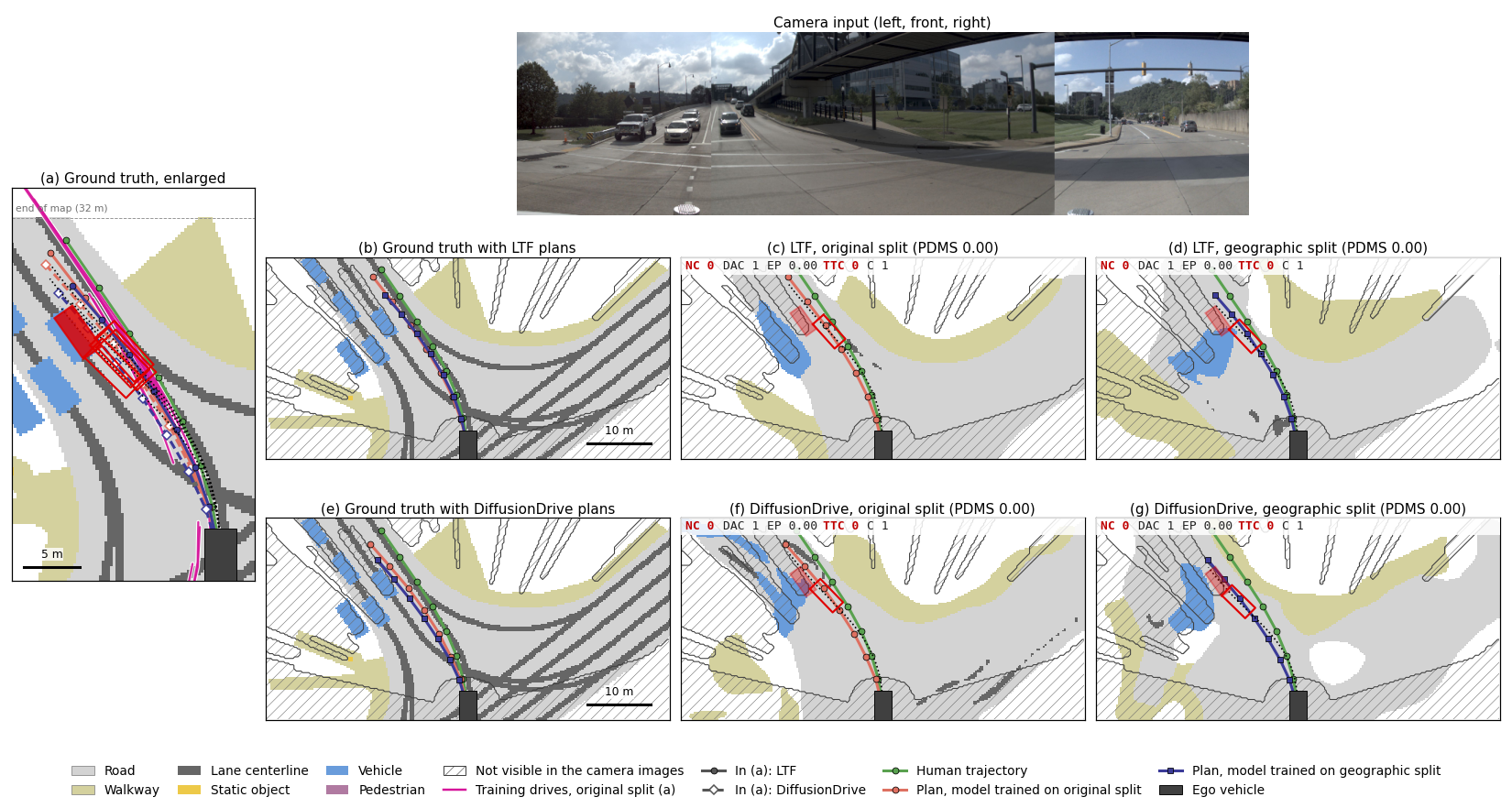}
  \caption{Pittsburgh, left turn with queued vehicles. All four models cut the turn tighter than the human
  driver and hit a queued vehicle (PDMS 0.00), regardless of the training split. Few training drives of
  the original split pass this place (a, pink).}\label{fig:app_qual_all_fail}
\end{figure*}

%% file: sec/tab/R16_full.tex
\begin{table*}[!htbp]
  \caption{PDMS gap $\text{PDMS}_\text{orig} - \text{PDMS}_\text{geo}$ on NSIT per city, at places with a training drive,
  without one, and their difference (\cref{tab:R16}), with 95\,\% intervals. Bold: the interval excludes zero. Las Vegas (grey) has only 41 scenes without a
  training drive.}\label{tab:R16_full}
  \centering
  \small
  \begin{tabular*}{\textwidth}{@{\extracolsep{\fill}}llrrccc@{}}
    \toprule
    Model & City & $n$ with & $n$ without & With training drive & Without & Difference \\
    \midrule
    DiffusionDrive & Singapore & 1{,}570 & 739 & $\mathbf{+6.5}$ {\scriptsize $[5.0, 8.5]$} & $+0.4$ {\scriptsize $[-0.9, 1.8]$} & $\mathbf{+6.1}$ {\scriptsize $[3.9, 8.9]$} \\
     & Boston & 593 & 182 & $\mathbf{+3.7}$ {\scriptsize $[1.5, 6.0]$} & $-1.6$ {\scriptsize $[-5.1, 1.7]$} & $\mathbf{+5.3}$ {\scriptsize $[1.5, 8.7]$} \\
     & Pittsburgh & 1{,}383 & 210 & $\mathbf{+16.1}$ {\scriptsize $[11.2, 21.6]$} & $+2.7$ {\scriptsize $[0.0, 5.1]$} & $\mathbf{+13.4}$ {\scriptsize $[7.9, 19.3]$} \\
     & \textcolor{gray}{Las Vegas} & \textcolor{gray}{3{,}223} & \textcolor{gray}{41} & \textcolor{gray}{$+3.8$ {\scriptsize $[2.5, 5.1]$}} & \textcolor{gray}{$+5.7$ {\scriptsize $[-1.1, 13.3]$}} & \textcolor{gray}{$-2.0$ {\scriptsize $[-9.7, 5.0]$}} \\
     & All & 6{,}769 & 1{,}172 & $\mathbf{+6.9}$ {\scriptsize $[5.3, 8.7]$} & $+0.7$ {\scriptsize $[-0.5, 1.9]$} & $\mathbf{+6.2}$ {\scriptsize $[4.3, 8.4]$} \\
    \midrule
    LTF & Singapore & 1{,}570 & 739 & $\mathbf{+3.4}$ {\scriptsize $[0.8, 5.9]$} & $\mathbf{-3.7}$ {\scriptsize $[-6.3, -1.8]$} & $\mathbf{+7.1}$ {\scriptsize $[4.4, 9.9]$} \\
     & Boston & 593 & 182 & $\mathbf{+5.5}$ {\scriptsize $[2.9, 8.4]$} & $-2.1$ {\scriptsize $[-6.9, 2.5]$} & $\mathbf{+7.6}$ {\scriptsize $[2.3, 13.4]$} \\
     & Pittsburgh & 1{,}383 & 210 & $\mathbf{+7.7}$ {\scriptsize $[4.8, 11.2]$} & $-1.0$ {\scriptsize $[-6.3, 3.0]$} & $\mathbf{+8.7}$ {\scriptsize $[4.0, 14.3]$} \\
     & \textcolor{gray}{Las Vegas} & \textcolor{gray}{3{,}223} & \textcolor{gray}{41} & \textcolor{gray}{$+4.4$ {\scriptsize $[2.9, 6.3]$}} & \textcolor{gray}{$+2.2$ {\scriptsize $[-6.9, 9.1]$}} & \textcolor{gray}{$+2.2$ {\scriptsize $[-5.3, 11.4]$}} \\
     & All & 6{,}769 & 1{,}172 & $\mathbf{+5.0}$ {\scriptsize $[3.7, 6.2]$} & $\mathbf{-2.8}$ {\scriptsize $[-4.8, -1.1]$} & $\mathbf{+7.7}$ {\scriptsize $[5.7, 9.8]$} \\
    \midrule
    DrivoR & Singapore & 1{,}570 & 739 & $\mathbf{+3.7}$ {\scriptsize $[2.1, 5.7]$} & $-0.1$ {\scriptsize $[-2.7, 1.9]$} & $\mathbf{+3.7}$ {\scriptsize $[1.2, 7.2]$} \\
     & Boston & 593 & 182 & $\mathbf{+4.0}$ {\scriptsize $[1.9, 6.3]$} & $+2.7$ {\scriptsize $[-0.4, 6.8]$} & $+1.2$ {\scriptsize $[-3.2, 5.0]$} \\
     & Pittsburgh & 1{,}383 & 210 & $\mathbf{+9.1}$ {\scriptsize $[5.3, 13.0]$} & $+1.4$ {\scriptsize $[-1.3, 5.2]$} & $\mathbf{+7.7}$ {\scriptsize $[2.6, 12.4]$} \\
     & \textcolor{gray}{Las Vegas} & \textcolor{gray}{3{,}223} & \textcolor{gray}{41} & \textcolor{gray}{$+1.9$ {\scriptsize $[1.2, 2.8]$}} & \textcolor{gray}{$-8.5$ {\scriptsize $[-18.5, 0.5]$}} & \textcolor{gray}{$+10.4$ {\scriptsize $[1.5, 20.4]$}} \\
     & All & 6{,}769 & 1{,}172 & $\mathbf{+4.0}$ {\scriptsize $[2.9, 5.2]$} & $+0.3$ {\scriptsize $[-1.5, 1.8]$} & $\mathbf{+3.6}$ {\scriptsize $[1.8, 5.6]$} \\
    \midrule
    Ego-MLP & Singapore & 1{,}570 & 739 & $\mathbf{-2.4}$ {\scriptsize $[-3.8, -1.2]$} & $-1.6$ {\scriptsize $[-4.5, 0.8]$} & $-0.8$ {\scriptsize $[-3.8, 2.8]$} \\
     & Boston & 593 & 182 & $+1.3$ {\scriptsize $[-1.0, 3.6]$} & $\mathbf{+4.8}$ {\scriptsize $[2.3, 7.3]$} & $\mathbf{-3.5}$ {\scriptsize $[-7.0, -0.4]$} \\
     & Pittsburgh & 1{,}383 & 210 & $\mathbf{+3.9}$ {\scriptsize $[2.0, 6.0]$} & $\mathbf{+5.3}$ {\scriptsize $[0.1, 12.3]$} & $-1.4$ {\scriptsize $[-8.6, 4.2]$} \\
     & \textcolor{gray}{Las Vegas} & \textcolor{gray}{3{,}223} & \textcolor{gray}{41} & \textcolor{gray}{$+2.7$ {\scriptsize $[1.7, 3.7]$}} & \textcolor{gray}{$+4.2$ {\scriptsize $[0.8, 8.1]$}} & \textcolor{gray}{$-1.5$ {\scriptsize $[-5.5, 2.1]$}} \\
     & All & 6{,}769 & 1{,}172 & $\mathbf{+1.6}$ {\scriptsize $[0.8, 2.5]$} & $+0.8$ {\scriptsize $[-1.4, 3.1]$} & $+0.8$ {\scriptsize $[-1.5, 3.0]$} \\
    \bottomrule
  \end{tabular*}
\end{table*}

%% file: sec/tab/tab_seed_robustness.tex
\begin{table*}[!htbp]
  \caption{NSIT PDMS gap $\text{PDMS}_\text{orig} - \text{PDMS}_\text{geo}$ for the seed mean (95\,\% interval)
  and its range over all 9 pairings of the three original-split and three geo-split training seeds, overall
  and at places with and without a training drive.}\label{tab:seed_robustness}
  \centering
  \small
  \setlength{\tabcolsep}{3pt}
  \begin{tabular*}{0.8\textwidth}{@{\extracolsep{\fill}}llccc@{}}
    \toprule
    Model & & NSIT & With drive & Without \\
    \midrule
    DiffusionDrive & seed mean & $+6.01$ {\scriptsize $[4.65, 7.58]$} & $+6.93$ {\scriptsize $[5.48, 8.74]$} & $+0.69$ {\scriptsize $[-0.46, 1.85]$} \\
     & range, 9 pairings & {\scriptsize $+5.39$ to $+6.55$} & {\scriptsize $+6.40$ to $+7.32$} & {\scriptsize $-0.44$ to $+2.17$} \\
    \midrule
    LTF & seed mean & $+3.82$ {\scriptsize $[2.64, 4.97]$} & $+4.96$ {\scriptsize $[3.74, 6.28]$} & $-2.76$ {\scriptsize $[-4.69, -1.10]$} \\
     & range, 9 pairings & {\scriptsize $+2.84$ to $+4.86$} & {\scriptsize $+3.68$ to $+6.36$} & {\scriptsize $-4.33$ to $-1.51$} \\
    \bottomrule
  \end{tabular*}
\end{table*}

%% file: sec/tab/tab_map_ablation.tex
\begin{table*}[!htbp]
  \caption{Map-loss ablation on NSIT, seed~0: PDMS with and without the map loss per training split, the effect
  of removing it (without $-$ with), and the resulting gaps (\cref{tab:R29}). Bold: the 95\,\% interval excludes zero.}\label{tab:map_ablation}
  \centering
  \small
  \begin{tabular*}{0.85\textwidth}{@{\extracolsep{\fill}}llccc@{}}
    \toprule
    Model & Training split & With map loss & Without & Without $-$ with \\
    \midrule
    DiffusionDrive & Original & 87.5 & 87.2 & $-0.3$ {\scriptsize $[-0.7, 0.1]$} \\
     & Geo & 81.8 & 81.7 & $-0.1$ {\scriptsize $[-0.9, 0.6]$} \\
     & Gap orig $-$ geo & $\mathbf{+5.6}$ {\scriptsize $[4.2, 7.2]$} & $\mathbf{+5.5}$ {\scriptsize $[4.1, 7.1]$} & $-0.1$ {\scriptsize $[-0.8, 0.5]$} \\
    \midrule
    LTF & Original & 80.3 & 78.3 & $\mathbf{-2.0}$ {\scriptsize $[-3.5, -0.6]$} \\
     & Geo & 76.8 & 72.5 & $\mathbf{-4.3}$ {\scriptsize $[-5.8, -3.1]$} \\
     & Gap orig $-$ geo & $\mathbf{+3.5}$ {\scriptsize $[1.9, 5.0]$} & $\mathbf{+5.8}$ {\scriptsize $[4.2, 7.4]$} & $\mathbf{+2.4}$ {\scriptsize $[0.3, 4.4]$} \\
    \bottomrule
  \end{tabular*}
\end{table*}

%% file: sec/tab/tab_valval.tex
\begin{table}[!htbp]
  \caption{PDMS on the 3,859 scenes (81 drives) in the validation set of both splits. Gap:
  $\text{PDMS}_\text{orig} - \text{PDMS}_\text{geo}$; over the 9 seed pairings, it ranges from $+6.6$ to $+11.1$
  (DiffusionDrive) and $+2.8$ to $+6.9$ (LTF). Difference: gap with minus
  gap without a training drive, weighted per city as in \cref{tab:R16}. Bold: the 95\,\% interval excludes zero.}\label{tab:valval}
  \centering
  \small
  \setlength{\tabcolsep}{3pt}
  \begin{tabular*}{\linewidth}{@{\extracolsep{\fill}}lcccc@{}}
    \toprule
    Model & Orig. & Geo & Gap & Difference \\
    \midrule
    DiffusionDrive & 83.6 & 74.9 & $\mathbf{+8.7}$ {\scriptsize $[6.6, 11.2]$} & $\mathbf{+5.9}$ {\scriptsize $[3.0, 9.4]$} \\
    LTF & 76.3 & 71.5 & $\mathbf{+4.8}$ {\scriptsize $[2.9, 6.8]$} & $\mathbf{+10.5}$ {\scriptsize $[7.6, 14.2]$} \\
    Ego-MLP & 53.4 & 51.2 & $\mathbf{+2.2}$ {\scriptsize $[0.5, 4.1]$} & $+0.6$ {\scriptsize $[-1.8, 2.5]$} \\
    \bottomrule
  \end{tabular*}
\end{table}

%% file: sec/tab/tab_pittsburgh_holdout.tex
\begin{table*}[!htbp]
  \caption{Models trained with vs.\ without Pittsburgh. A: NAVSIMv1 PDMS on navtest, Pittsburgh (22 drives) and the
  other cities (114 drives). B: NAVSIMv2 EPDMS, single stage with non-reactive traffic, on the 450 recorded
  first-stage scenes of navhard (174 in Pittsburgh, 19 drives). Gap: with $-$ without, with 95\,\% interval;
  bold if it excludes zero. For DiffusionDrive, 24 navtest scenes (9 in Pittsburgh) are dropped on both sides
  because one run returned an invalid trajectory there.}\label{tab:pittsburgh_holdout}
  \centering
  \small
  \setlength{\tabcolsep}{4pt}
  \begin{tabular*}{\textwidth}{@{\extracolsep{\fill}}lcccccccc@{}}
    \toprule
    & \multicolumn{4}{c}{A: navtest, PDMS} & \multicolumn{4}{c}{B: navhard stage 1, EPDMS} \\
    \cmidrule(lr){2-5} \cmidrule(l){6-9}
    & \multicolumn{3}{c}{Pittsburgh} & Other & \multicolumn{3}{c}{Pittsburgh} & Other \\
    \cmidrule(lr){2-4} \cmidrule(lr){5-5} \cmidrule(lr){6-8} \cmidrule(l){9-9}
    Model & With & Without & Gap & Gap & With & Without & Gap & Gap \\
    \midrule
    DiffusionDrive & 85.8 & 75.0 & $\mathbf{+10.8}$ {\scriptsize $[7.7, 15.0]$} & $\mathbf{+0.5}$ {\scriptsize $[0.2, 0.8]$} & 68.3 & 48.9 & $\mathbf{+19.4}$ {\scriptsize $[12.0, 28.9]$} & $\mathbf{+2.0}$ {\scriptsize $[0.2, 4.0]$} \\
    LTF & 79.4 & 68.8 & $\mathbf{+10.6}$ {\scriptsize $[7.7, 13.6]$} & $\mathbf{+0.6}$ {\scriptsize $[0.1, 1.1]$} & 62.3 & 46.4 & $\mathbf{+15.9}$ {\scriptsize $[9.8, 22.7]$} & $+0.8$ {\scriptsize $[-2.1, 3.9]$} \\
    Ego-MLP & 55.4 & 52.1 & $\mathbf{+3.3}$ {\scriptsize $[2.1, 5.0]$} & $+0.4$ {\scriptsize $[-0.1, 1.0]$} & 38.1 & 35.2 & $+2.9$ {\scriptsize $[-2.7, 10.0]$} & $+2.3$ {\scriptsize $[-0.1, 4.8]$} \\
    \bottomrule
  \end{tabular*}
\end{table*}